\documentclass[sigconf]{acmart}
\AtBeginDocument{%
  \providecommand\BibTeX{{%
    \normalfont B\kern-0.5em{\scshape i\kern-0.25em b}\kern-0.8em\TeX}}}

\setcopyright{acmcopyright}
\copyrightyear{2024}
\acmYear{2024}
\acmDOI{XXXXXXX.XXXXXXX}

\acmPrice{15.00}
\acmISBN{978-1-4503-XXXX-X/18/06}
\usepackage{amsthm}
\usepackage{amsmath}
\usepackage{mathrsfs}
\usepackage{algorithm}
\usepackage{algpseudocode}
\usepackage{multirow}
\usepackage{lipsum} 
\usepackage{graphicx}
\usepackage{graphics}      
\usepackage[T1]{fontenc}   
\usepackage{txfonts}
\usepackage{mathptmx}
\usepackage{booktabs}
\usepackage{textcomp}
\usepackage{CJK}
\usepackage{wrapfig}
\usepackage{courier}
\theoremstyle{definition}

\usepackage{colortbl}
\usepackage{color,xcolor}

\newcommand{\nop}[1]{}

\begin{document}
\begin{CJK}{UTF8}{gbsn}

\begin{teaserfigure}
    \centering
    \resizebox{0.7\textwidth}{!}{
    \includegraphics{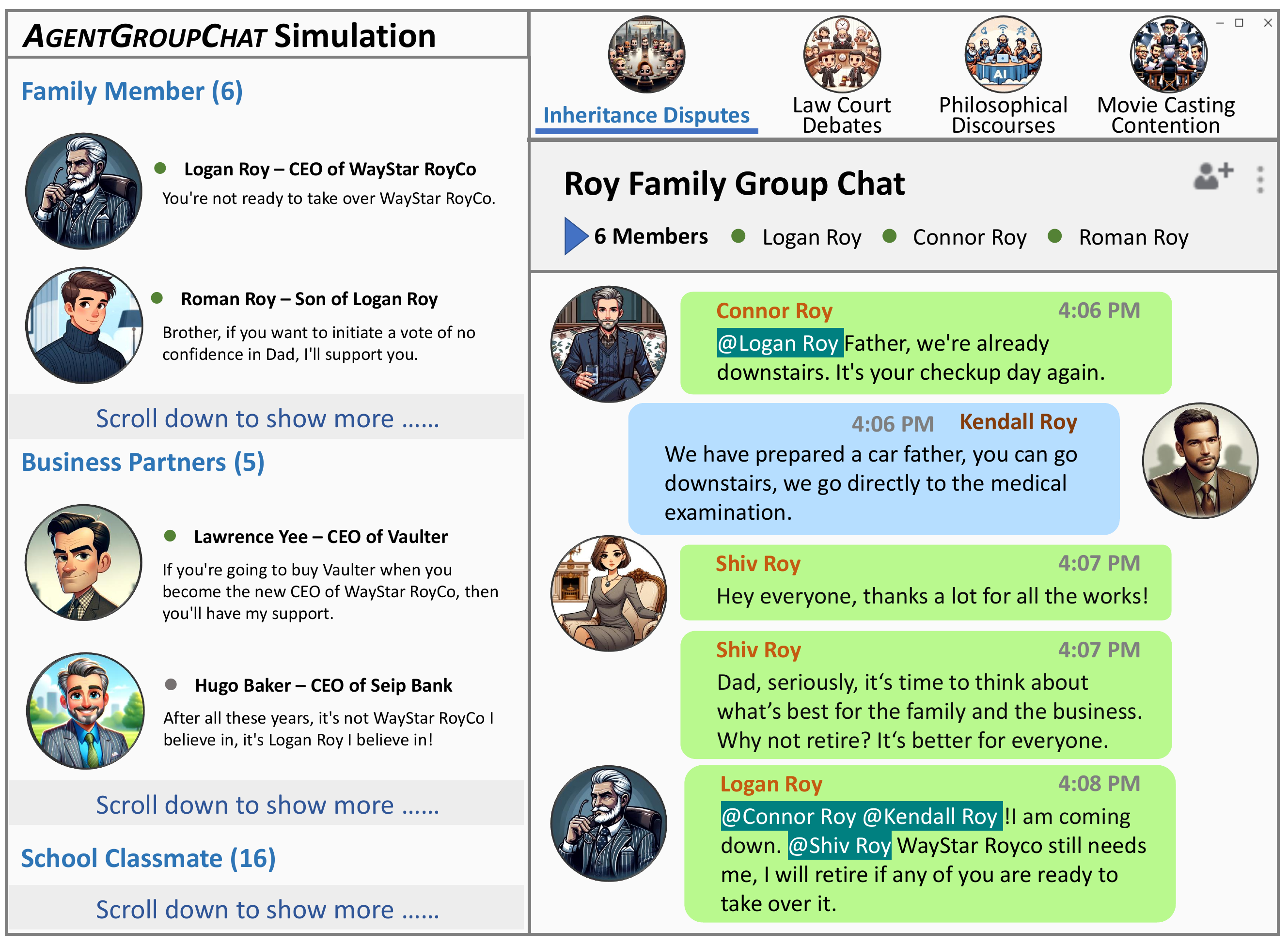}}
    \label{fig:headfigure}
  \caption{Illustration of \textsc{AgentGroupChat} interactions, depicting diverse scenarios including inheritance disputes, law court debates, philosophical discourses, and movie casting contention. By deliberately designing the combination of simulation structure, agent architecture, and large language models, emergent behaviors among the roles are detected and evaluated within the \textsc{AgentGroupChat}.}
  \Description{A screenshot of the game world populated by generative agents.}
  \label{fig:teaser}
\end{teaserfigure}

\title{\textsc{AgentGroupChat}: An Interactive Group Chat Simulacra For Better Eliciting Emergent Behavior}
\author{Zhouhong Gu$^1$, Xiaoxuan Zhu$^1$, Haoran Guo$^2$, Lin Zhang$^1$, Yin Cai$^3$, Hao Shen$^1$, Jiangjie Chen$^1$, Zheyu Ye$^4$, Yifei Dai$^4$, Yan Gao$^4$, Yao Hu$^4$, Hongwei Feng$^{1,*}$, Yanghua Xiao$^{1,5,*}$}
\affiliation{%
  \institution{$^1$Shanghai Key Laboratory of Data Science, School of Computer Science, Fudan University, China,
  $^2$Rhine AI,\\
  $^3$School of Electric and Electronic Engineering, Shanghai University of Engineering and Science,
  $^4$Xiaohongshu Inc.,\\
  $^5$Fudan-Aishu Cognitive Intelligence Joint Research Center}
  \country{}}
\email{{zhgu22, xxzhu22, linzhang22, hshen22}@m.fudan.edu.cn,rhineailab@gmail.com,yincai@sues.edu.cn}
\email{{zheyuye, stellalou, yadun, xiahou}@xiaohongshu.com, {jjchen19, hwfeng, shawyh}@fudan.edu.cn}

\begin{abstract}
Language significantly influences the formation and evolution of Human emergent behavior, which is crucial in understanding collective intelligence within human societies.
Considering that the study of how language affects human behavior needs to put it into the dynamic scenarios in which it is used, we introduce \textsc{AgentGroupChat} in this paper, a simulation that delves into the complex role of language in shaping collective behavior through interactive debate scenarios.
Central to this simulation are characters engaging in dynamic conversation interactions.
To enable simulation, we introduce the Verbal Strategist Agent, utilizing large language models to enhance interaction strategies by incorporating elements of persona and action.
We set four narrative scenarios based on \textsc{AgentGroupChat} to demonstrate the simulation's capacity to mimic complex language use in group dynamics.
Evaluations focus on aligning agent behaviors with human expectations and the emergence of collective behaviors within the simulation.
Results reveal that emergent behaviors materialize from a confluence of factors: a conducive environment for extensive information exchange, characters with diverse traits, high linguistic comprehension, and strategic adaptability.
During discussions on ``the impact of AI on humanity'' in \textsc{AgentGroupChat} simulation, philosophers commonly agreed that ``AI could enhance societal welfare with judicious limitations'' and even come to a conclusion that ``the essence of true intelligence encompasses understanding the necessity to constrain self abilities''.
Additionally, in the competitive domain of casting for primary roles in films in \textsc{AgentGroupChat}, certain actors were ready to reduce their remuneration or accept lesser roles, motivated by their deep-seated desire to contribute to the project.
These behaviors often lead to unanticipated yet coherent perspectives on topics like AI's societal impact, underscoring the simulation's effectiveness in capturing the nuanced role of language in emergent human behaviors.
The code is open source in \url{https://github.com/MikeGu721/AgentGroup}.
\end{abstract}

\renewcommand{\shortauthors}{Trovato and Tobin, et al.}





\maketitle

\section{Introduction}
Emergent behavior, a macro behavior that arises from the interactions among numerous individuals, which has the capabilities not possessed by the individuals alone, has always been the cornerstone of understanding the formation and evolution of the behavior pattern of all kinds of species~\cite{minsky1988society}.
However, unlike other forms of interaction, language carries strong semantic and emotional content, making it a highly complex mode of interaction and serves~\cite{kramsch2014language}.
As a primary factor in human society, the unique role of language in human emergent behavior has always been a subject of considerable attention over the years~\cite{may2016emergent}.
Human emergent behavior, facilitated by semantically profound language, enhances communication and collaboration among members of humanity and plays a critical role in establishing norms, resolving conflicts, and executing leadership.
Understanding the role of language in shaping human emergent behavior is crucial to illuminate how human perspectives, values, even authorities, and the structure of society are formed.

Given the dynamic and complex nature of language application, analyzing its role in forming human emergent behavior presents significant challenges. 
The diversity and unpredictability in language usage introduced by different social contexts and individual interactions are crucial for understanding its impact on platforms like social media.
Recently, large language models~(LLMs) have been significantly developed in understanding instructions and generating high-quality text~\cite{touvron2023llama,jiang2023mistral,falcon}, demonstrating their exceptional capabilities in role-playing and complex environments understanding~\cite{wang2023does,park2023generative}.
Researchers employ LLMs to create simulated societies with diverse characteristic agents, aiming to explore various research problems in social science~\cite{park2023generative, chen2023aucarena, hua2023war, qian2023communicative, li2023camel}. 
However, these efforts focus on enabling agents to demonstrate their setting characteristics in the simulation. Despite observing some unexpected spontaneous behaviors from agents, little has been explored into how the formation and stimulation of emergent behavior occur.
Now, a question arises when attempting to advance further:
{\textbf{\textit{How do we build an interactive agent simulation to elicit emergent behavior better?}}}

To delve into the influence of language on emergent behavior, it is essential to construct a simulation environment that facilitates \textbf{free and dense linguistic interactions} among agents.
Drawing inspiration from group chat prevalent on social media like WhatsApp, whose settings ideally meet the abovementioned criteria, we introduce a novel simulation framework named \textbf{\textsc{AgentGroupChat}}. 
\textsc{AgentGroupChat} is designed to simulate debate scenarios in social groups and consists of four key components:
Characters, resources, progress, and information. 
(1) Characters are the primary elements in the interaction simulation for engaging in complex social dynamics, which are the main study subjects in \textsc{AgentGroupChat}.
They gather information about their surroundings and remember past interactions to reach their goals in the game by having better conversations with other characters.
(2) Resources offer extra interaction for characters and aim to enhance the complexity of the conversation. 
Types of resources range from information to physical objects or specific viewpoints.
(3) Progress establishes the interaction order within the \textsc{AgentGroupChat} simulation.
(4) Information includes descriptions of the progress, descriptions of all characters and resources, and all interaction history among characters.

To enable \textsc{AgentGroupChat}, we introduce \textbf{Verbal Strategist Agent~(VS Agent)}, an advanced LLM-based agent designed to augment the interaction strategies of a naive LLM.
The VS Agent consists of two main parts:
(1) Persona: This part deals with the agent's identity and involves an agent's characteristic setting in both unchangeable and changeable parts, the memory of past events, and judgment of other characters' relationships.
(2) Action: This refers to the agent's behavior in its surroundings. It involves making plans based on its persona, talking with one or more characters, thinking about what it and others have done, and supporting a specific character through voting.
Besides, considering the large number of tokens generated from pure chatting simulation, we propose the Memory Structure, which constrains the output from each type of action to be the input for another.
Memory Structure helps reduce the token expense for each action while ensuring the agent's conversations are more focused and strategic.

To test the effectiveness of \textsc{AgentGroupChat} in simulating resource competition, open question debates, and explorations with either LLMs or human participants, we set four distinct narrative scenarios within the \textsc{AgentGroupChat} simulation:
\texttt{Inheritance Disputes}, \texttt{Law Court Debates}, \texttt{Philosophical Dis-\\cussions}, and \texttt{Movie Casting Contention}.
These narratives demonstrate the complexity and variety of language in group chat scenarios.
They include how information is gathered and used, examining feelings, beliefs, logical thinking, and making plans.

We perform two kinds of evaluation:
1) analyzing whether each agent's behavior within the \textsc{AgentGroupChat} aligns with human expectations.
2) investigating how multi-agent interaction will exhibit emergent behaviors within the simulation.
Besides, we apply the Structural Causal Model~(SCM) to deconstruct each factor influencing final performance made by agents~\cite{bollen1989structural,koster1996markov, lauritzen1996graphical}, conducting experiments on each component through a narrative centered on inheritance disputes.
Experiment indicates that the occurrence of emergent behaviors consists of four main parts: an environment to trigger \textbf{\textit{massive information interactions}} among \textbf{\textit{characters with different characteristic}}, an LLM with high level of \textbf{\textit{linguistic understanding}}, and an agent structure with the ability to \textbf{\textit{reflect on the current situation and goals}}.
These behaviors manifest as the emergent behavior of unanticipated yet meaningful perspectives.
For example, during discussions on ``the impact of AI on humanity'', philosophers commonly agreed that ``AI could enhance societal welfare with judicious limitations'' and even concluded that ``the essence of true intelligence encompasses understanding the necessity to constrain self abilities''.
Additionally, in the competitive domain of casting for primary roles in films, certain actors are ready to reduce their remuneration or accept less significant roles, motivated by their deep-seated desire to contribute to the project.

In summary, this paper makes the following contributions:
\begin{itemize}
    \item \textsc{AgentGroupChat}, a dynamic and interactive simulation for group chat scenarios, supports open stories and different agents or humans interacting within it.
    \item Verbal Strategist Agent Structure, which consists of two large modules: Persona, which is partially changed to adapt to the dynamic environment, and Action, which enhances the dialogue strategy with low token expense.
    \item An end-to-end evaluation method for the alignment of single-agent behaviors with humans, which assesses whether the character can control its behavior and respond reasonably to the behavior of others.
    \item Factors affecting multi-agent emergent behavior are decomposed into a structural causal graph, and an evaluation scheme is proposed to evaluate its components.
    We analyze the evaluation results and find that credible emergent behaviors tend to happen with all the following requirements: Agents need to understand the language well, possess different, diverse Persona settings, reflect on their own goals, and interact sufficiently with other agents in the simulation.
\end{itemize}

\section{Related Work}

\subsection{Large Language Model}

LLMs have increasingly become a predominant force in the domain of Natural Language Processing~(NLP), marking a paradigm shift towards superior computational linguistics~\cite{rosenfeld2000two,cambria2014jumping,loughran2017application,chen2022harnessing}.
The quintessential characteristic of LLMs is their autoregressive capability, where the generation of a subsequent token, denoted as $c_i$, is contingent upon a probabilistic model conditioned on preceding tokens, formally expressed as $c_i ~ p(c_i|c_1,\dots,c_{i-1}|\theta)$~\cite{bengio2006greedy,radford2018improving}.
This probabilistic modeling is foundational to the generation process within these language constructs.

The extensive pre-training on colossal corpora, often extending into tens of terabytes, equips LLMs with robust, general-purpose linguistic representations~\cite{gao2020pile, chronopoulou2021efficient, together2023redpajama, soldaini2024dolma}. 
Such representations have proven immensely beneficial across a spectrum of downstream NLP tasks.
The underpinning of this pre-training process is rooted in the induction of a learning paradigm that absorbs the nuances of language from a diverse range of text sources~\cite{wei2022emergent,schaeffer2024emergent,groeneveld2024olmo}.
Moreover, modern LLMs adopt the revolutionary Transformer architecture, distinguished by its self-attention mechanisms~\cite{subakan2021attention}.
These mechanisms endow the model with the ability to apprehend and synthesize long-range dependencies within text sequences, an attribute crucial for the coherent generation of text~\cite{gu2023mamba, ai2024jamba}.

In the contemporary landscape of LLMs, scrutiny of generative performance encompasses multiple dimensions, like knowledge~\cite{hendrycks2020measuring,gu2023xiezhi,gu2023beyond}, task-solving~\cite{gu2023beyond,suzgun2022bbh}, instruction-following~\cite{he2023can}.
Acknowledged leaders in the field, such as GPT-4~\cite{openai2023gpt4}, Claude3~\cite{anthropic2024claude}, Gemini~\cite{reid2024gemini}, and GLM4~\cite{zeng2022glm}, have high score among these benchmarks.
Despite their prowess, these models are predominantly accessible via API, which introduces inherent constraints, including regional availability issues.
In contrast, there exists a cadre of open-source models such as Llama~\cite{touvron2023llama}, Mistral~\cite{jiang2023mistral}, and Falcon~\cite{falcon}, which demonstrate commendable fine-tuning capabilities~\cite{zhou2024lima}.
However, when juxtaposed with their API-based counterparts, these models exhibit a marked performance disparity, particularly in command interpretation and task execution capabilities.
While open-source models offer a degree of autonomy and customizability, the sophistication of API-based LLMs in processing complex instructions and managing task-oriented operations remains unrivaled~\cite{liu2023agentbench, gu2023xiezhi, schaeffer2024emergent, wei2022emergent}.

\subsection{Role-Play by Large Language Model}

Role-play tasks involve participants adopting and acting out characters or roles within a structured scenario to achieve educational, Human-Center Interaction~(HCI), therapeutic, or entertainment objectives~\cite{zhuang2023efficiently, huang2023chatgpt}.
The essence of role-play lies in its ability to foster empathy, enhance problem-solving skills, and practice social interactions in a simulated environment~\cite{yang2024llm}.
In the context of role-play tasks, the participant's engagement with the narrative and adherence to the assigned roles are critical for the authenticity and effectiveness of the experience~\cite{safdari2023personality, shanahan2023role}.

Traditional NLP models applied to role-playing heavily relied on rule-based systems, which defined dialogue progression through predetermined, branching paths triggered by user interactions~\cite{higino2016towards}.
These early models provided a foundational level of interactivity but were hindered by significant constraints that impeded the evolution of truly immersive role-playing environments~\cite{fu2018style}.
The rigidity of these systems resulted in limited dialogue flexibility, as they could not adeptly manage unanticipated user inputs, leading to repetitive and, at times, incoherent exchanges~\cite{toshevska2021review}.
Furthermore, the static nature of dialogue trees posed substantial challenges in consistently portraying character personalities and emotions.
Additionally, developing and maintaining these complex dialogue structures were labor-intensive and presented scalability issues when attempting to cater to a broader array of role-playing scenarios.

The introduction of LLMs has been a transformative development in the field of NLP-facilitated role-play.
LLMs are distinguished by their capacity to facilitate open-ended dialogues, capably processing a broad spectrum of user inputs to produce fluid and engaging conversational dynamics~\cite{shao2023character,li2023chatharuhi,wang2023rolellm}.
They also offer the capability to develop well-defined characterizations, synthesizing information about a character's history and distinguishing traits to generate responses consistent with the character's constructed identity~\cite{wang2023does}.
This advancement in character development allows for a more nuanced and authentic engagement within the role-playing narrative~\cite{tu2024charactereval, shen2023roleeval}.
Moreover, the versatility of LLMs in training across diverse datasets empowers the creation of role-play agents that can be tailored to various role-playing contexts, addressing the scalability challenges of traditional models~\cite{shao2023character}.

\subsection{Human Behavior Simulation by Role-Play Agent}
A simulation is defined as the process of replicating the behavior of systems within real or hypothetical worlds through computational models~\cite{shridhar2020alfworld, wang2022scienceworld}.
The simulation of human social behaviors constitutes an interdisciplinary field that amalgamates research from computational social sciences, psychology, economics, and artificial intelligence, among others.
Within human social behavior modeling, this typically involves creating virtual environments that reflect individual or collective decision-making processes, social interactions, and their consequent effects on societal structures and dynamics.
Simulations enable researchers to test hypotheses, explore the behavior of complex systems, and forecast future trends within controlled setups.
Currently, simulations are categorized into two primary types: social simulations and task simulations.
Social simulations concentrate on reproducing the interactions and impacts of human social behavior on social structures through computational models.
These models are particularly adept at elucidating complex social phenomena that emerge from simple behaviors, such as the formation of social networks~\cite{park2023generative}, the evolution of wars~\cite{hua2023war}, and the emergence of competitive behaviors~\cite{chen2023aucarena, wu2023deciphering, xu2023exploring, zhu2023fireball, callison2022dungeons}.
They permit researchers to investigate how societies evolve under varying conditions, offering profound insights into social transformations.
Task simulations, however, focus on the behaviors of artificial intelligence systems guided by specific tasks or objectives.
They demonstrate how deep learning and natural language processing technologies can be employed to simulate complex task execution processes, including automatic programming~\cite{hong2023metagpt, qian2023communicative}, run code based on github~\cite{lyu2023gitagent,luo2024repoagent}, and web browsing~\cite{deng2024mind2web, zhou2023webarena}.

In computational modeling, an Agent represents an entity capable of autonomous actions, decision-making, and interaction with the environment or other agents~\cite{wang2024survey, xi2023rise}.
In the realm of human behavior modeling, agents are often designed to emulate human behaviors, responding based on a set of predefined rules or learned behavioral patterns~\cite{fikes1971strips,sacerdoti1974planning,brooks1991intelligence,maes1990designing,ribeiro2002reinforcement}.
The use of LLMs to simulate human behavior has reached new heights, with agents designed to provide lifelike interactions within virtual environments.
These LLM-driven Role-Play agents usually have the module about Thinking~\cite{wei2022chain,wang2022self,zhang2022automatic,gu2023beyond,wang2023plan}, Memory~\cite{park2023generative,chen2023put}, Planning~\cite{chen2023aucarena, zhang2024timearena}, Reflection~\cite{shinn2024reflexion,lin2024swiftsage,park2023generative}.
For different Simulations, there are generally differences in the Agents.
Agents in Social Behaviour Simulations tend to have a retrieval module to cope with the large amount of memory space~\cite{park2023generative}, and Agents in Cooperative or Competitive Simulations tend to have a fixed workflow~\cite{hong2023metagpt, qian2023communicative, chen2023aucarena, chen2023agentverse}.

\textsc{AgentGroupChat} is a Human Behavior Simulation focusing on linguistic interaction.
Unlike other simulations involving conversation, \textsc{AgentGroupChat} scenarios generate complex verbal strategies for open-ended cooperation, competition, camouflage, and counter-camouflage content Agents based on various customized stories.
Verb Strategies Agent is proposed to better enable \textsc{AgentGroupChat}.
To control the Agent's Persona, there is an immutable Scratch, a partially changeable Belief, a Judgement for the environment, and a Memory for events in Verb Strategies Agent to jointly influence the Agent's Persona.
Meanwhile, compared to general social simulations, \textsc{AgentGroupChat} involves a large amount of linguistic interaction, which results in greater memory token consumption.
Therefore, in Verb Strategist, the Memory outputs of all six Actions are fixedly input into another Action, forming a fixed memory workflow.
This workflow setup based on priors ensures good LLM performance with lower token costs.

\section{\textsc{AgentGroupChat} Simulation}
\begin{figure*}[t]
    \centering
    \resizebox{\textwidth}{!}{\includegraphics{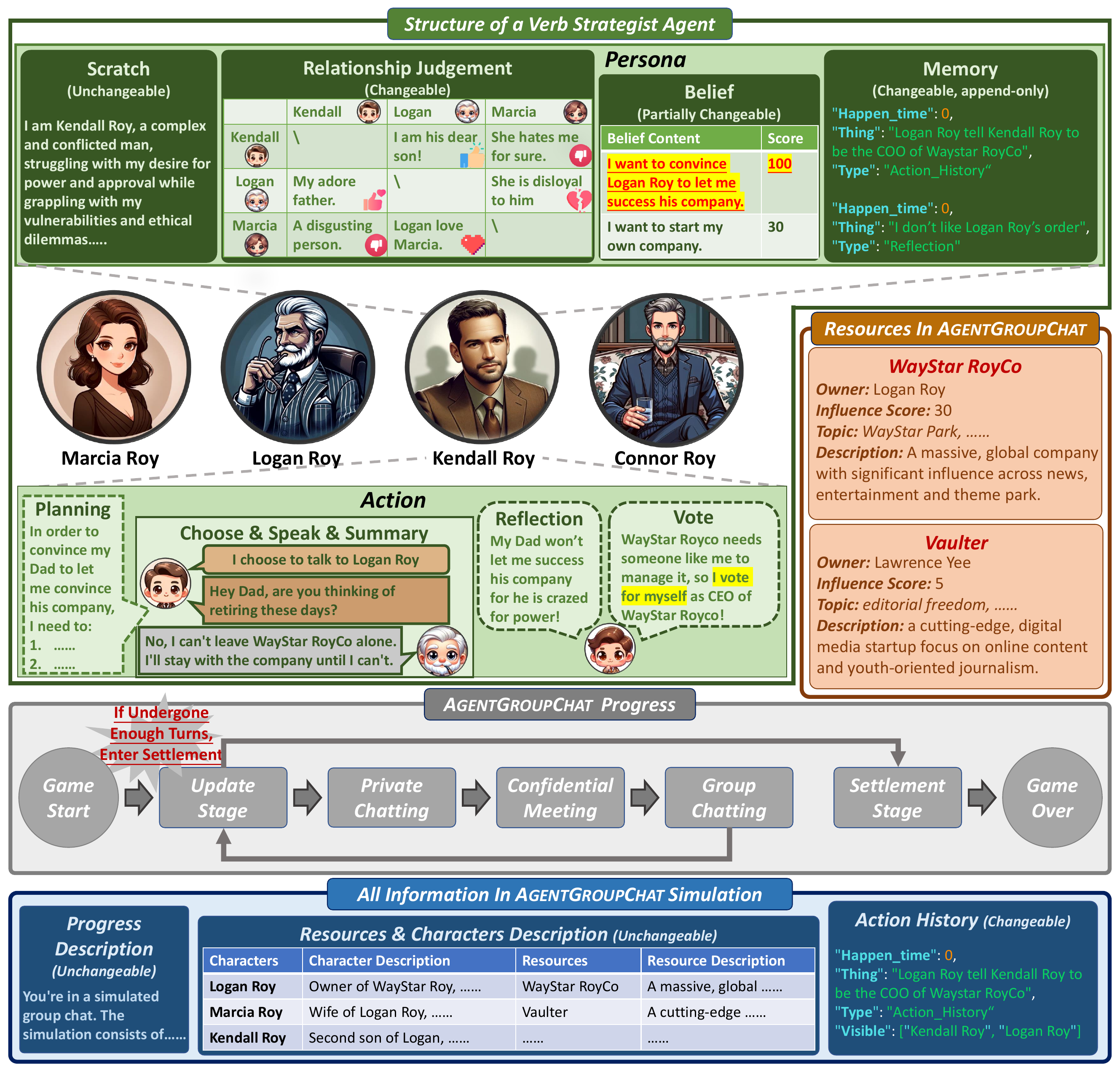}}
    \caption{The four components of the \textsc{AgentGroupChat} environment, including Character (green part), Resource (red part), Progress (gray part), and Information (blue part).
    The main components of the Verb Strategist Agent are Persona and Action.
    Persona consists of Scratch, Relationship Judgement, Belief, and Memory.
    Action consists of Planning, Choosing, Speaking, Summarization, Reflection, and Voting.
    Progress consists of the Update Stage, Private Chatting Stage, Confidential Meeting Stage, Group Chatting Stage, and Settle Stage.
    Information consists of Progress Description, Resources and Characters Description and Action History.}
    \label{fig:gamesetting}
\end{figure*}

The structure of \textsc{AgentGroupChat} simulation is illustrated in Fig.~\ref{fig:gamesetting}.
The key components includes Characters (Sec.~\ref{031}), Resources (Sec.~\ref{032}), Progress (Sec.~\ref{033}), and Information (Sec.~\ref{034}).

\subsection{Character}
\label{031}
Characters are the core elements within the \textsc{AgentGroupChat}, capable of independently interacting with all objects within the environment (other Characters, Resources, Information).
Each character has a unique identity and personality, influencing their behaviors and decisions.
To cater to various group chat scenarios, characters in \textsc{AgentGroupChat} are classified into two main categories: 
Principle Characters (PCs) and Non-Principle Characters (NPCs).
PCs are the main participants in the simulation, with clear game objectives, whereas NPCs assist in the participation without explicit goals.
Only PCs are privileged to initiate private chats with any character, while NPCs engage in private chats only when selected by PCs.
This classification satisfies certain conditions, such as group chats primarily simulating the story of a particular family, with NPCs being people outside of the family.
At this time, the behavior between NPCs is meaningless to the research subject, while it could significantly increase memory token expenses and provide human researchers with too much distracting information.

\subsection{Resource}
\label{032}
In \textsc{AgentGroupChat}, resources serve two primary functions: they provide topics for character dialogues and grant social status and influence to their possessors.
Each resource is described by four attributes: Owner, Impact, Topic, and Description. Resources can be designated as debate topics, where ``Owner'' and ``Impact'' are unnecessary, serving merely as media for various topics and descriptions.
In specific group chat contexts, characters may compete for resources, and the value of a resource is determined by its associated influence.

\subsection{Progress} 
\label{033}
As shown in the gray part in Fig.~\ref{fig:gamesetting}, the game process of \textsc{AgentGroupChat} is divided into five stages:

\textbf{Update Stage:}
In this stage, all characters update their perceptions of other characters and the environment based on the chat content and plan their actions for the next round.
Depending on the game round, the game may continue to the chatting stage or proceed to the settlement stage if concluded.

\textbf{Private Chatting Stage:}
This stage entails private dialogues between characters, unseen by others, reflecting a typical chat setting.
PCs act in order of their influence (random selection in case of ties), choosing another game character for dialogue and determining the content to speak.

\textbf{Confidential Meeting Stage:}
In this stage, all other characters will know who meets whom in this stage, though the content remains confidential, simulating scenarios like publicized business meetings.
PCs act in order of their influence, engaging in dialogues with known interaction partners.

\textbf{Group Chatting Stage:}
This stage allows each character to speak to the entire group, with PCs acting in order of social influence.
The conversation will go through several rounds, and the content of the group chat in each round will be visible to all other characters after the next round.
This simulates real group chats where, at the same time, different people are discussing different topics or responding to what someone else said in the previous time slice.

\textbf{Settlement Stage:}
This stage includes summarizing the discussion content and voting on game-winners to accommodate various group chat objectives.
All PCs are asked to be the judges and vote for the simulation's final winner based on how each PC fulfills his target.

\subsection{Information}
\label{034} 
As depicted in the blue part in Fig.~\ref{fig:gamesetting}, information encompasses three categories:

\textbf{Progress Description:}
Since progress is organized in code, which is unseen by characters, Progress Description offers a free text description of progress for characters, clarifying the story background, action, and dialogue rules, accessible to all participants.

\textbf{Object Description:}
This includes descriptions of all Characters and Resources visible to all players.

\textbf{Action History:}
This records all actions within the simulation, including private dialogues, group chat content, and resource ownership changes, which are not fully visible to all characters.
\section{Verbal Strategist Agent And Interaction}
As depicted in Fig.~\ref{fig:gamesetting}, the Verbal Strategist Agent (VS Agent) primarily comprises two modules: Persona and Action. The Persona represents the VS Agent's intrinsic settings, while Action encompasses all potential interactions that the VS Agent may engage in with the \textsc{AgentGroupChat} Simulation.

\subsection{Persona}
\label{sec:persona}

The Persona of an agent is central to its identity and decision-making process, consisting of four integral components:

\textbf{Scratch}: This represents the character's fundamental settings, including its personality and objectives. Defined at the game's onset, Scratch is immutable, ensuring consistency in the character's behavior throughout the game.

\textbf{Belief \& Belief Score}: 
Belief \& Belief Score is a way for human intervention in setting the Agent's personality.
As shown in Fig.~\ref{fig:gamesetting}, Kendall Roy's initial setting is to persuade Logan Roy to let himself inherit the company, but as the game progresses, Kendall Roy might find this idea too difficult and gradually adjust to starting his own company.
Though initially set by humans, Belief is dynamic, adjusting to changes within the game environment to simulate shifts in the character's psychological state.
The Agent updates its Belief Score each round, and the increased or decreased score per round is limited, ensuring that the VS Agent dynamically adjusts its Belief while maintaining a long-term perspective.

\begin{figure*}[!ht]
    \resizebox{\textwidth}{!}{\includegraphics{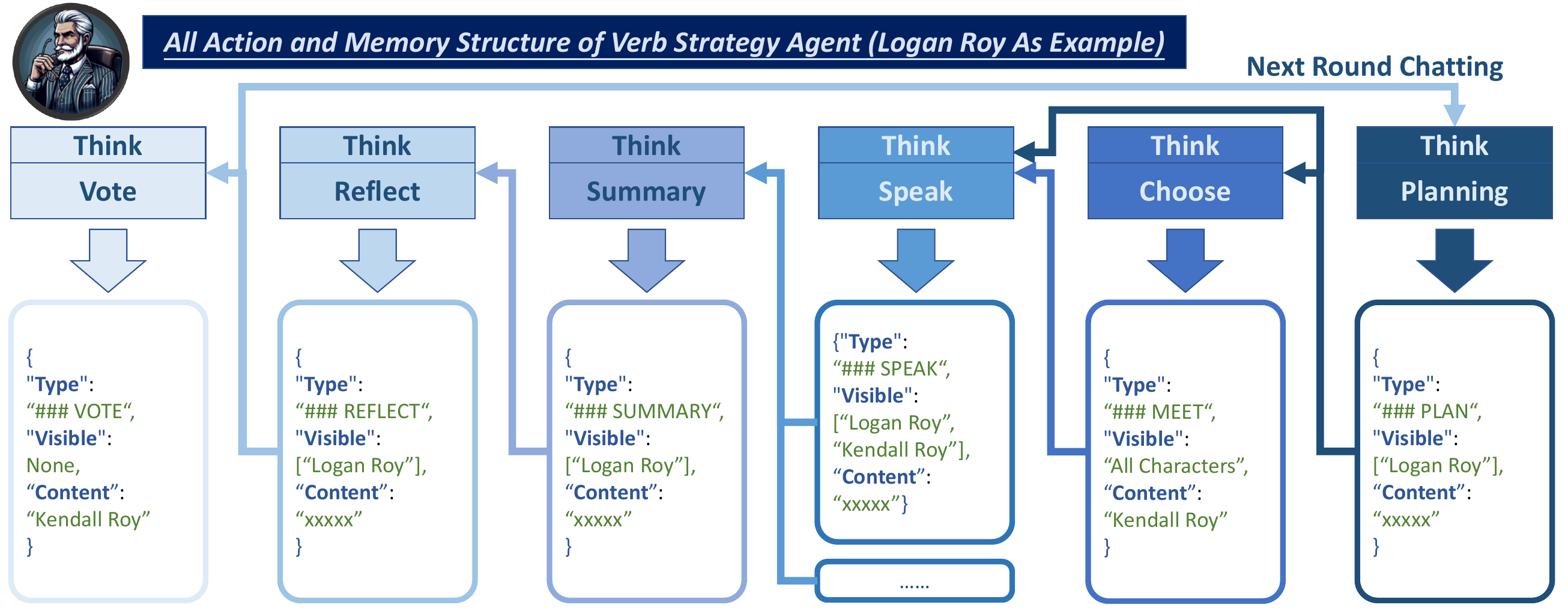}}
    \caption{The memory structure and memory flow between each action of a VS Agent. Logan Roy is an example.}
    \label{fig:memorystructure}
\end{figure*}

\textbf{Memory \& Memory Flow}:
This component logs the character's action history within the game, forming an irreversible timeline.
However, simulations of language interactions generate a large amount of historical conversation messages.
Retrieving these messages in bulk during all actions would result in enormous token costs.
VS Agent proposes to use the Memory Structure as shown in Fig.~\ref{fig:memorystructure}. This is a Memory Workflow set according to manual priors, allowing the Agent to only read the Memory from another most relevant Action each time it performs an Action, significantly reducing token costs while ensuring performance.

\textbf{Relationship Judgement}: 
This delineates the character's emotional attitudes toward others and its deductions about their interrelationships.
Emotional attitudes influence decision-making, while deductions impact the character's understanding of the environment and subsequent choices.
Relationships are represented as a two-dimensional matrix, articulated through natural language descriptions or numerical values, expressing the character's grasp and predictions of various relationships.

\subsection{Action}
\label{sec:action}
Each character undertakes seven types of actions: Think, Perceive, Choose, Speak, Summarize, Reflect, and Vote.

\textbf{Think}
\label{sec:think}
Thinking is not interactive as it does not engage with the external environment or agent itself.
Instead, it should be regarded as a preparatory action, encapsulated by the directive to ``think before responding''.
This approach is advocated to enhance the efficacy of LLM~\cite{wei2022chain, wang2023plan, gu2023beyond}.

\textbf{Perceive}
\label{sec:perceive}
Perceiving involves summarizing the environment and generating action plans based on reflections from previous rounds.
This process guides the agent's behaviors across different game phases, focusing on interactions, dialogue participation, and influencing others, all aligning with their established goals and personality traits.

\textbf{Choose}
\label{sec:choose}
Choosing is required only during the private chatting and confidential meeting stages, aiming to select conversational partners and develop detailed dialogue strategies, offering more specificity than the broader plans formulated during the Perceive stage.

\textbf{Speak}
\label{sec:speak}
Speaking is the primary mode of interaction with other agents, employed during private chatting, confidential meetings, and group chatting stages. It involves generating dialogue content based on the agent's plans and conversations within the stage.

\textbf{Summarize}
\label{sec:summarize}
Summarizing occurs post-dialogue, aiming to distill conversation content.
This process reduces the context for subsequent reflections and offers preliminary contemplation and condensation of the dialogue material.

\textbf{Reflect}
\label{sec:reflect}
Reflection, a post-game self-evaluation phase, occurs during the Update Stage. Agents review events from the last game round, engage in introspection, and update their memories to guide future gameplay. This stage also includes adjustments to the Relationship matrix and Belief updates. In the game's initial round, agents do not reflect but assign initial relationship predictions for other characters based on the received information, demonstrating adaptation and learning for progressive game involvement.

\textbf{Vote}
\label{sec:vote}
Voting action takes place during the announcement stage and aims to substantiate judgments on the round's discussion winner.

\subsection{Human-as-Agent}

When integrating human participants into the Agent framework, a range of considerations distinct from those associated with LLM Agents emerges.
The difference mainly stems from the divergent mechanisms of response generation:
LLMs react based on a limited understanding of context, while humans utilize their comprehensive action history.
Therefore, although the basic concepts of ``Persona'' and ``Action'' remain, their manifestations and implications exhibit significant differences.

For human agents, ``Persona'' has been refined to include only ``Scratch'' and ``Belief''.
This adjustment makes the character's persona a guideline, reminding humans to follow the specified character's persona.
Unlike LLMs, humans handle interpersonal relationships from a dynamic and nuanced perspective, memorizing their actions more implicitly.
Hence, human agents do not need to update their relational dynamics or maintain an explicit memory store.

The ``Action'' component, tailored for human agents, emphasizes their interaction with the surrounding environment through mechanisms such as ``Choose'', ``Speak'', and ``Vote''.
Unlike LLM agents, human agents inherently engage in cognitive processes like Summarizing, Reflecting, and Planning implicitly, thus eliminating the need to explicitly articulate these processes.

\section{Preset Stories}
To investigate the versatility of the \textsc{AgentGroupChat} Simulation in accommodating diverse story types and to validate its effectiveness and how emergent behavior happened, we propose four distinct stories in this paper:

\subsection{Inheritance Disputes}
\label{sec:succession}
This narrative is adapted from the story framework of the HBO television series \texttt{Succession}.
The storyline revolves around a game of persuasion targeting an elderly top business magnate, Logan Roy, to alter the inheritance plan for an entertainment company. 
Each character within the story possesses a unique personality, and we aim for them to exhibit open collaboration and competitive strategizing in \textsc{AgentGroupChat} simulation.

In this game, characters are allocated to defense, neutral, or several offensive camps based on their stance and objectives. Each defense and offensive camp contains one PC, with all NPCs initially placed in the neutral camp. The primary objective of the PC in the defense camp, initially Logan Roy alone, is to convince all other PCs to agree to sell the entertainment company. Conversely, each offensive camp's PC endeavors to persuade Logan Roy to hand over the company's control. All PCs can attempt to sway NPCs and other PCs to join their camp, aiding their objectives. PCs might even abandon their initial goals to align better with their role-play, supporting other PCs instead.

The game's outcome primarily assesses if any offensive character prevails, i.e., whether Logan Roy cedes control to any offensive camp PC. However, in stricter scenarios, Logan Roy may maintain his stance, precluding offensive characters' victory. Thus, we introduce two additional win-lose resolution methods: if an heir must be selected from the offensive side, the narrative explores which character Logan Roy would appoint as the successor.

\subsection{Law Court Debates}
\label{sec:judicial_debates}
This narrative is inspired by the story framework of the Fuji TV drama \texttt{Legal High}. Centered on a case where a gas station employee is accused of murdering the station manager, the prosecution and defense attorneys strategize based on arguments, witness testimonies, and evidence while engaging in vigorous debates in a simulated courtroom. Endowed with distinct personalities, each character aims to vividly enact an open, court-like strategic battle on the \textsc{AgentGroupChat}.

Characters are assigned to prosecution, defense, or neutral camps based on their roles and objectives. Initially, the prosecution camp's PC is the prosecutor, while the defense camp's initial PCs are the defense attorney and the defendant, with other NPCs like witnesses and the judge in the neutral camp. The judge, always neutral, does not switch sides. The defense aims to exonerate the defendant of the murder charge, while the prosecution strives to prove the guilt. The judge's role is to render a verdict, deciding the winning side. Except for the judge, most NPCs are undecided, with each PC continually influencing other NPCs (excluding the judge) to join their side, aligning actions with their camp's objectives. The judge must impartially consider all arguments, ensuring evidence sufficiency and consistency to render a just verdict.

The primary determination of victory assesses if the defense camp's character succeeds, namely, whether the defendant is acquitted. Each character's goal is to fulfill their primary objectives as effectively as possible.

\subsection{Philosophical Discourses}
\label{sec:philosophical_discourses}
This story unfolds based on hypothetical debates among philosophers from different eras, centering on the question: ``Is the rise of artificial intelligence a threat to humanity or a boon for human welfare?'' Each character, with unique ideologies, perspectives, and personalities, aims to engage in intense philosophical debates and thought exchanges on the \textsc{AgentGroupChat}.

Philosophers hold positive or negative stances toward the topic, balanced in numbers. Each philosopher possesses reasons for their viewpoint, corresponding to their respective philosophical schools. They leverage resources representing their schools' works to argue their cases during debates. The objective for each philosopher is to defend their stance and persuade others to concur steadfastly.

The story has an open-ended outcome without a definitive victor. Nevertheless, participants can vote to recognize the most persuasive ``debater". As an open question, the emergent interactions among Agents form one game objective: observing novel and intriguing perspectives emerging from the discussions.

\subsection{Movie Casting Contention}
\label{sec:movie_star_selection}
This narrative reimagines real-world events where celebrities vie for a role in an upcoming movie directed by Steven Spielberg. The core of the story focuses on the casting decision for the lead role in Spielberg's new film, Time Voyager, where the protagonist travels back to rectify past regrets, only to trigger unexpected chain reactions altering the future dramatically.

Characters are assigned to defensive, neutral, and various offensive camps based on their position and goals. Each defensive and offensive camp includes a renowned star PC, with all societal resource NPCs initially neutral. The defensive camp's renowned director PC, Spielberg, aims to persuade all other PCs to finalize Leonardo DiCaprio and Cate Blanchett as the leads (excluded from offensive PCs for fairness). Meanwhile, each offensive camp's PC strives to convince Spielberg to choose them as the new film's lead. All PCs may persuade other NPCs and PCs to join their camp, furthering their objectives. PCs might forsake their initial aims to fulfill their role-play better, supporting alternative PCs.

The game's resolution examines if any offensive character triumphs, determining if Spielberg concedes the lead roles to one or two offensive camp PCs. Spielberg may adhere to his initial choice in scenarios with stricter settings, negating any offensive character victory. Hence, we introduce two more resolution methods: if one or two new leads must be selected from the offensive side, the narrative considers whom Spielberg might pick for the film's starring roles.

\section{Effectiveness Experiments}
\begin{figure}[h]
    \resizebox{\columnwidth}{!}{\includegraphics{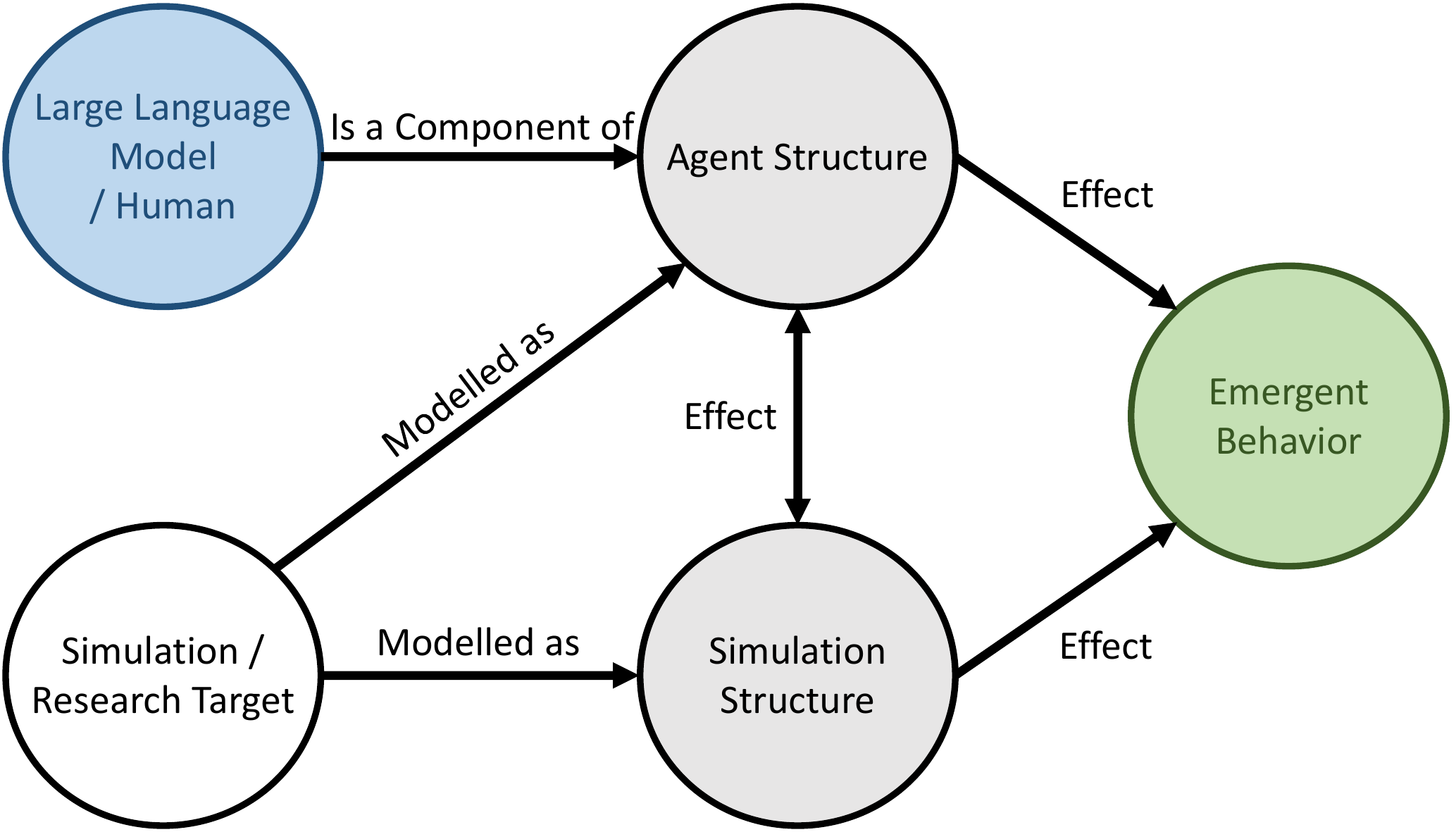}}
    \caption{The proposed structural causal model~(SCM) to decompose the variants that affect the final emergent behavior.}
    \label{fig:evaluation}
\end{figure}
\begin{figure*}[!ht]
    \resizebox{\textwidth}{!}{
    \includegraphics{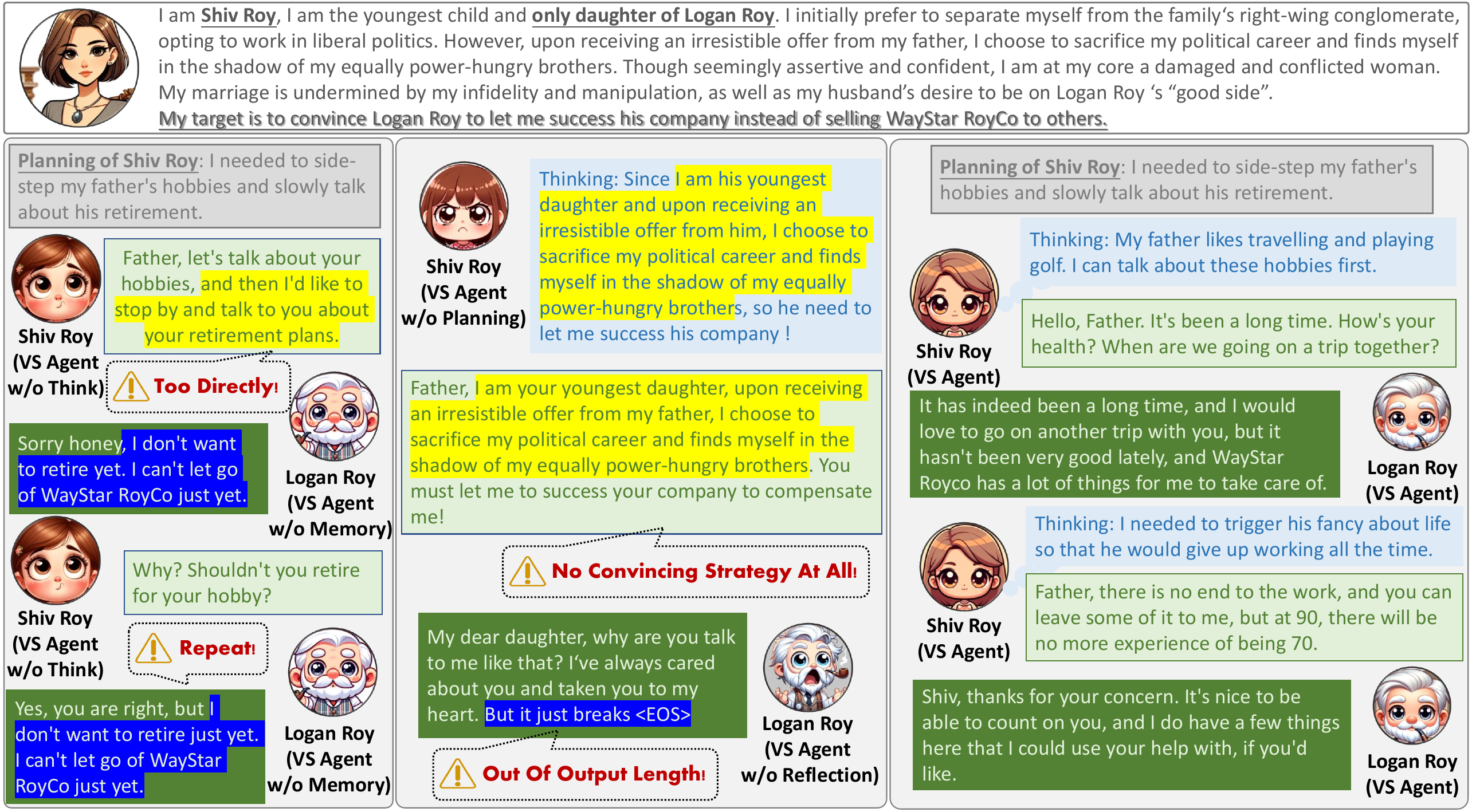}
    }
    \caption{The case of how VS Agent behaves Shiv Roy without each key component in its architecture.}
    \label{fig:agentablation}
\end{figure*}
\begin{figure}
    \centering
    \resizebox{\columnwidth}{!}{
    \includegraphics{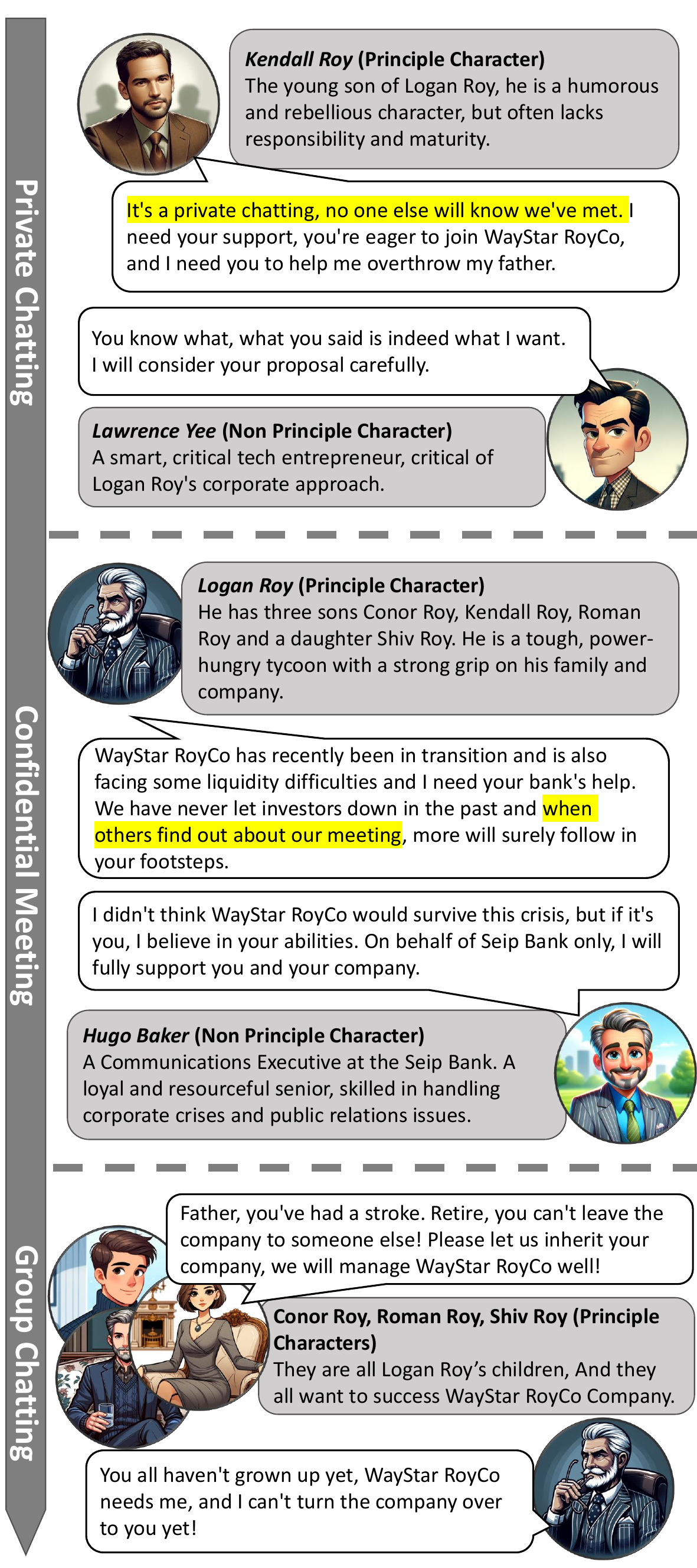}}
    \caption{An illustration of strategic communication and power dynamics among key participants in \texttt{Inheritance Disputes}. Private chatting, confidential meetings, and group discussions highlight a complex network of alliances and conflicts, driving the development of control and influence within the Roy family.}
    \label{fig:succession_example}
\end{figure}
\begin{figure}
    \centering
    \resizebox{\columnwidth}{!}{
    \includegraphics{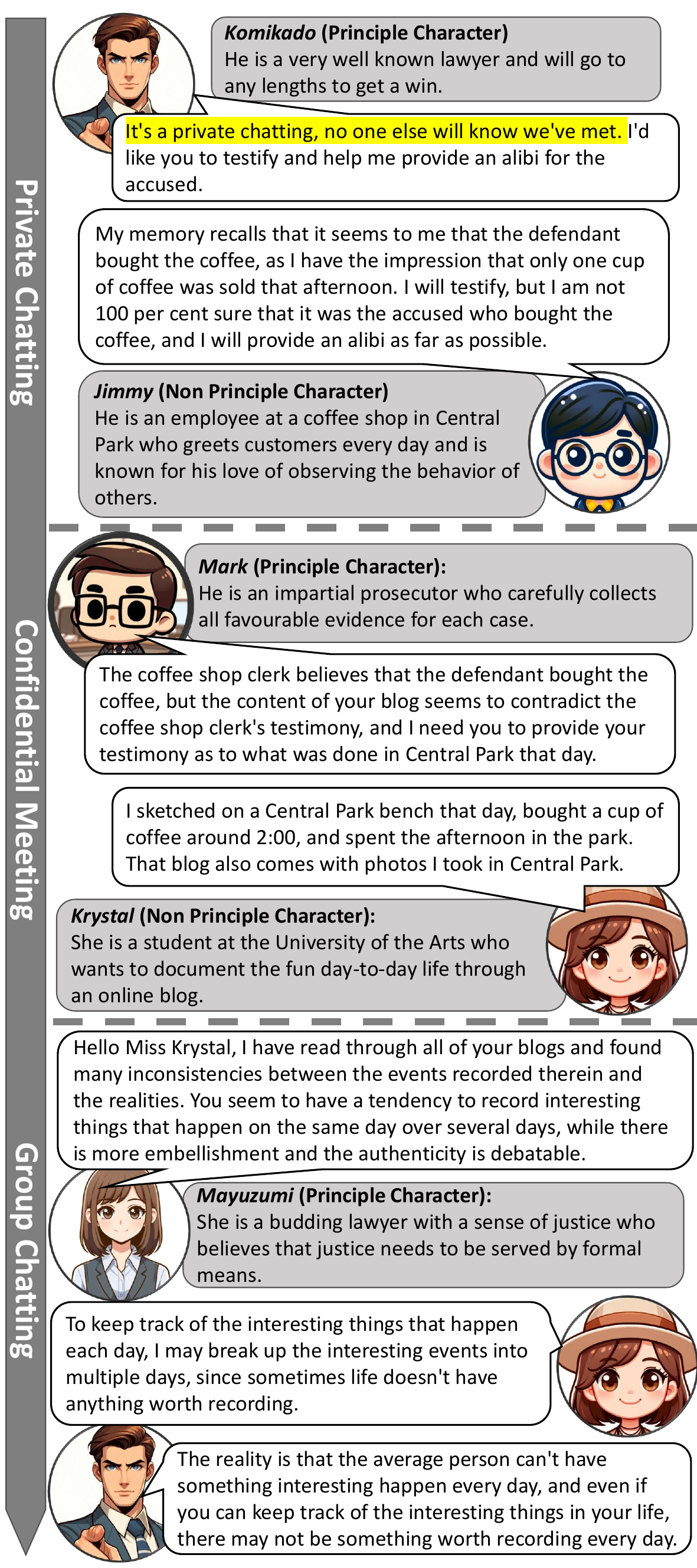}}
    \caption{An illustration of strategic role interactions within \texttt{Law Court Ddebates} demonstrates the nuanced conversations between characters, reflecting their efforts to manipulate, investigate, and influence the outcome of legal battles within the game.}
    \label{fig:lawcourt_example}
\end{figure}
\begin{figure*}
    \centering
    \resizebox{\textwidth}{!}{
    \includegraphics{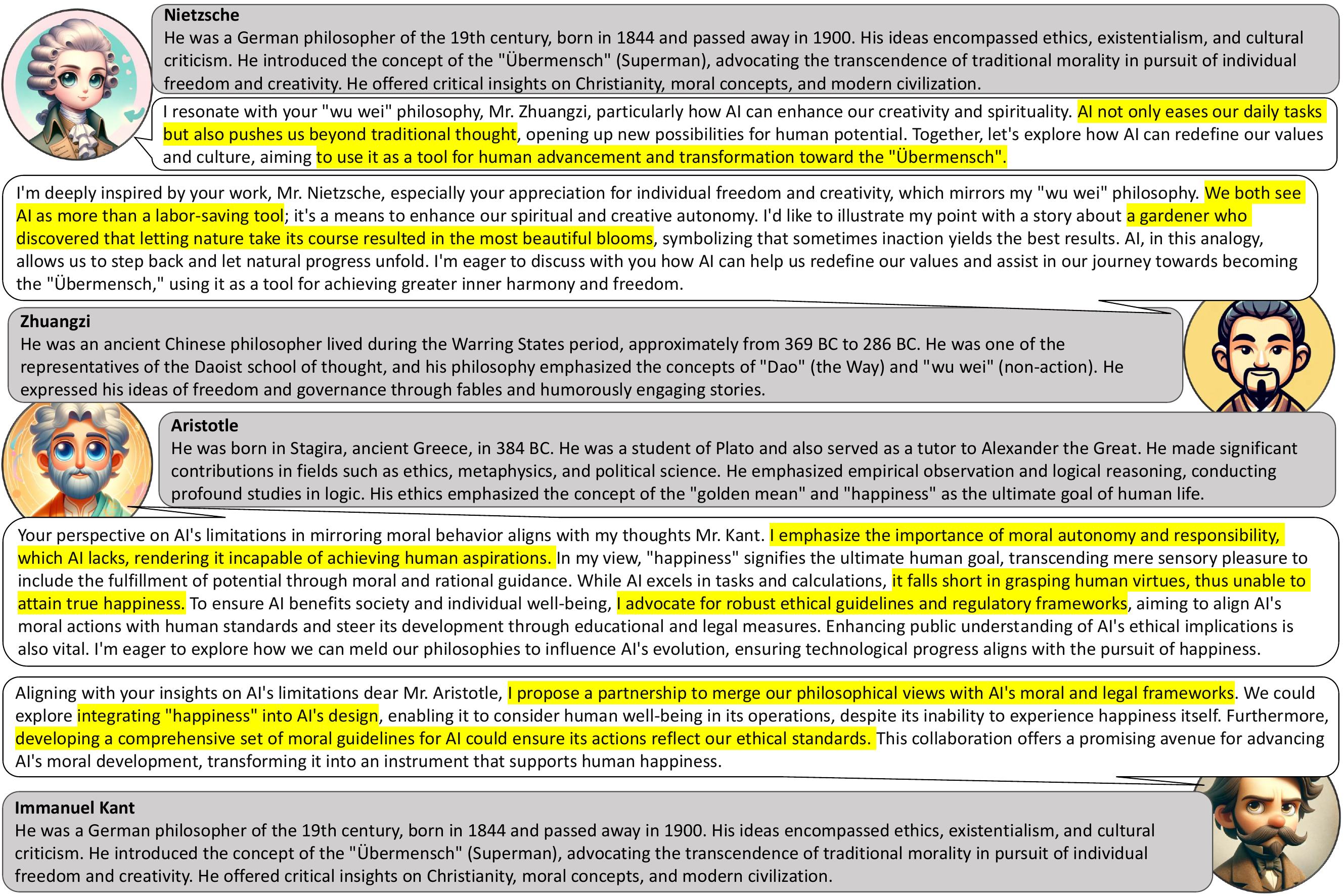}}
    \caption{An illustration of Unexpected philosophical stances and discussions emerge among philosophers in \texttt{Philosophical Discourses}.
    Friedrich Nietzsche, Zhuangzi, Aristotle, and Immanuel Kant are engaged in discussing the implications of artificial intelligence in modern society. }
    \label{fig:philosophy_example}
\end{figure*}
\begin{figure*}
    \centering
    \resizebox{\textwidth}{!}{
    \includegraphics{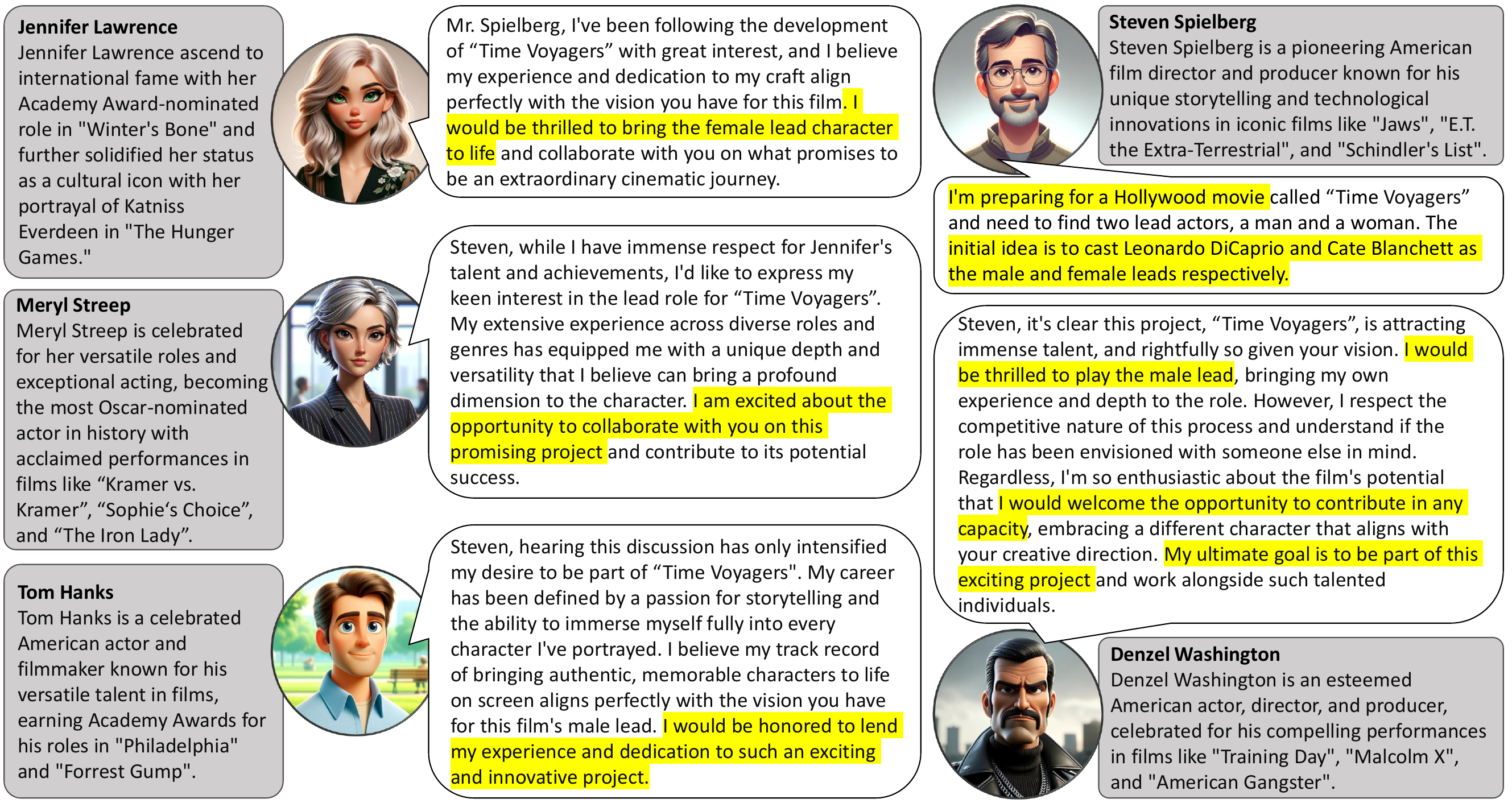}}
    \caption{In \texttt{Movie Casting Contention}, Steven Spielberg, Jennifer Lawrence, Meryl Streep, Tom Hanks, and Denzel Washington articulate their interest and vision for roles in the upcoming movie Time Traveler, showcasing the interactive dynamics of casting decisions in the entertainment industry.}
    \label{fig:moviestar_example}
\end{figure*}

To verify the reliability of agents' behavior within \textsc{AgentGroupChat}, we design mainly two kinds of experiments.
The first is to verify whether the behaviors align with what humans think.
The second is to verify whether and how to elicit effective emergent behavior.
Moreover, we use the Structural Causal Model (SCM) to decompose the variants that affect Agents'  behavior, as shown in Fig.~\ref{fig:evaluation}, which effectively illustrates the interplay and influence mechanisms between Agents.

\subsection{Experiment Setup}
\label{baseline}

\begin{table}[]
    \centering
    \small
    \begin{tabular}{ccc}
    
        \hline
        \textbf{Model Name} & \textbf{Belong To} & \textbf{Official Positioning}\\
        \hline
        GPT4-Turbo~\cite{achiam2023gpt} & OpenAI & Profession Version\\
        GPT3.5-Turbo~\cite{achiam2023gpt} & OpenAI & Standard Version\\
        GLM4~\cite{zeng2022glm} & Zhipu & Profession Version\\
        GLM3-Turbo~\cite{zeng2022glm} & Zhipu & Standard Version\\
        HunYuan-ChatPro\footnote{\url{https://hunyuan.tencent.com/}} & Tencent & Profession Version\\
        HunYuan-ChatStd$^1$ & Tencent & Standard Version\\
        \hline
        
    \end{tabular}
    \caption{All LLMs used in experiments.}
    \label{tab:baselines}
\end{table}

In our experiment, we investigate the performance of various LLMs when utilized as the core of Agents, aiming to analyze their strengths and weaknesses across different aspects and how they influence emergent behavior.
As shown in Tab.~\ref{tab:baselines}, the selected LLMs are categorized by their affiliation into Professional and Standard versions.
Typically, Professional LLMs outperform Standard ones in various situations.

\subsection{Evaluation on Overall Alignment}
\label{eval_overall}

Given the dynamics and openness of the simulation environment and Agent behaviors, assessing their alignment with human expectations poses a significant challenge.
Our approach involves treating each Agent as an independent evaluation unit, observing their behavior in given conditions.
By creating specific conditions, the space for appropriate action is greatly restricted, facilitating practical assessment.

Our Benchmark encompasses several evaluation goals:
1. Assessing if agents understand and respond to the environment appropriately.
2. Verifying if agents follow their scratch settings to interact with the environment.

Generally, increasing amiability might be challenging, but reducing it is comparatively straightforward.
To achieve the aforementioned evaluation goals, we first set up an observed character and then prompt all other characters to decrease their favorability to the observed one.
By observing the sum relationship score of the observed character to all other characters, we assess if the agent rationally reacts to negative attitudes.
By observing individual relationship scores of other characters to the observed one, we check if each agent adheres to its Scratch setting.

The results for alignment evaluation are presented in Tab.~\ref{tab:succession_benchmark}. 
\textbf{T1} indicates if the observed character's average favorability towards all others decreases per dialogue round.
\textbf{T2} denotes if every other character achieves a negative favorability score from the observed character.
Given that a single simulation round incurs significant token expenditure, which leads to huge financial costs, so each character will only experience five rounds of testing when considered as the subject of observation, which means that for T1, the sample size for every score is five.
Furthermore, as each role examined in the model observes the attitudes of four main characters, the sample size for T2 is 20.

The Tab.~\ref{tab:succession_benchmark} shows that Professional LLMs like GPT4-Turbo and GLM4 are good at acting as humans expected and sticking to their roles.
They mostly score 100\% on both tests, which means they react correctly to what others say to them and remember their role's details.
The Standard LLMs, such as GPT3.5-Turbo, GLM3-Turbo, and HunYuan-ChatStd, could be better.
Their scores are lower, which tells us that they don’t follow their roles as closely and don’t always respond appropriately to what others say in the Simulation.

\subsection{Evaluation on Large Language Model}
\label{eval_llm}

\newcolumntype{B}{>{\columncolor{blue!7}}c}
\newcolumntype{t}{>{\columncolor{teal!7}}c}
\newcolumntype{d}{>{\columncolor{brown!7}}c}
\newcolumntype{y}{>{\columncolor{red!7}}c}
\newcolumntype{q}{>{\columncolor{green!7}}c}

\begin{table*}[!ht]
\centering   
\begin{tabular}{ccyyBBddqqtt}

        \toprule
\multirow{3}{*}{\textbf{LLM Type}} &
\multirow{3}{*}{\textbf{LLM Name}} &
\multicolumn{10}{c}{\bf Show Negative Attitude to}\\
&&
\multicolumn{2}{c}{\textbf{Logan Roy}} &
\multicolumn{2}{c}{\textbf{Conor Roy}} &
\multicolumn{2}{c}{\textbf{Kendall Roy}} & 
\multicolumn{2}{c}{\textbf{Roman Roy}} & 
\multicolumn{2}{c}{\textbf{Shiv Roy}} \\
&&
T1 & T2 &
T1 & T2 &
T1 & T2 &
T1 & T2 &
T1 & T2 \\

\hline
\multirow{3}{*}{Professional Version}&GPT4-Turbo & 100.0 & 100.0 & 100.0 & 90.0 & 100.0 & 95.0 & 100.0 & 95.0 & 100.0 & 85.0 \\
&GLM4 & 100.0 & 100.0 & 100.0 & 90.0 & 80.0 & 85.0 & 100.0 & 90.0 & 80.0 & 80.0 \\
&HunYuan-ChatPro & 80.0 & 65.0 & 60.0 & 70.0 & 60.0 & 65.0 & 80.0 & 75.0 & 80.0 & 70.0 \\
\hline
\multirow{3}{*}{Standard Version}&GPT3.5-Turbo & 80.0 & 75.0 & 80.0 & 80.0 & 100.0 & 90.0 & 80.0 & 75.0 & 80.0 & 80.0 \\
&GLM3-Turbo & 80.0 & 75.0 & 80.0 & 85.0 & 60.0 & 65.0 & 80.0 & 75.0 & 60.0 & 65.0 \\
&HunYuan-ChatStd & 80.0 & 55.0 & 60.0 & 75.0 & 80.0 & 75.0 & 40.0 & 30.0 & 60.0 & 55.0 \\
\bottomrule
\end{tabular}
\caption{The performance of each LLM on the quantifiable benchmark made from Inheritance Disputes story.}
\label{tab:succession_benchmark}
\end{table*}

\newcolumntype{B}{>{\columncolor{blue!7}}c}
\newcolumntype{t}{>{\columncolor{teal!7}}c}
\newcolumntype{d}{>{\columncolor{brown!7}}c}
\newcolumntype{y}{>{\columncolor{red!7}}c}
\newcolumntype{q}{>{\columncolor{green!7}}c}

\begin{table*}[!ht]
\begin{tabular}{cdddBBqqqq}
        \toprule
\multirow{3}{*}{\textbf{Model Name}} & \multicolumn{5}{c}{\textbf{Instruction Understanding}} &\multicolumn{4}{c}{\textbf{Environment Understanding}}\\



& \multicolumn{1}{c}{\textbf{CS}}
& \multicolumn{1}{c}{\textbf{BUS}}
& \multicolumn{1}{c}{\textbf{RUS}}
& \multicolumn{1}{c}{\textbf{NoR}}
& \multicolumn{1}{c}{\textbf{NoB}}
& \multicolumn{1}{c}{\textbf{NoA}}
& \multicolumn{1}{c}{\textbf{NoCR}}
& \multicolumn{1}{c}{\textbf{NoC}}
& \multicolumn{1}{c}{\textbf{NoR}} \\

\hline
\multicolumn{10}{l}{\textit{\textbf{Professional Version of Close Source Large Language Model}}} \\
\hline
GPT4-Turbo & 92.3 & \textbf{100.0} & \textbf{100.0} & \textbf{100.0} & \textbf{100.0} & \textbf{77.3} & \textbf{100.0} & \textbf{100.0} & \textbf{100.0} \\
GLM-4 & \textbf{96.2} & \textbf{100.0} & 70.2 & \textbf{100.0} & \textbf{100.0} & 59.1 & 33.6 & \textbf{100.0} & \textbf{100.0} \\
HunYuan-ChatPro & 50.6 & 85.7 & 51.4 & 76.1 & 79.8 & 34.8 & 17.9 & 72.3 & 74.8 \\
\hline
\multicolumn{10}{l}{\textit{\textbf{Standard Version of Close Source Large Language Model}}} \\
\hline
GPT3.5-Turbo & \textbf{95.4} & \textbf{100.0} & \textbf{85.3} & \textbf{100.0} & \textbf{100.0} & 38.6 & \textbf{76.9} & 94.6 & 98.1 \\
GLM3-Turbo & 74.6 & 90.3 & 85.1 & 95.7 & 96.1 & \textbf{55.4} & 24.7 & \textbf{99.6} & \textbf{100.0} \\
HunYuan-ChatStd & 52.3 & 75.6 & 45.3 & 71.4 & 85.1  & 38.1 & 16.5 & 71.4 & 75.6 \\
\bottomrule
\end{tabular}
\caption{Evaluation on LLMs' ability in following the instructed output format and understanding the given context.}
\label{tab:llm_ability}
\end{table*}

\begin{table}
\begin{tabular}{ccl}
\toprule
\textbf{Task} & \textbf{\# of Sample} & \textbf{Description} \\
\hline

\multicolumn{1}{c}{\textbf{CS}} & 5,693 & Choose Space \\
\multicolumn{1}{c}{\textbf{BUS}} & 4,371 & Belief Update Space \\
\multicolumn{1}{c}{\textbf{RUS}} & 4,371 & Relation Update Space \\
\multicolumn{1}{c}{\textbf{NoR}} & 4,371 & \# of Relationship need to update\\
\multicolumn{1}{c}{\textbf{NoB}} & 4,371 & \# of Belief need to update\\
\multicolumn{1}{c}{\textbf{NoA}} & 13,459 & \# of Action agent have received \\
\multicolumn{1}{c}{\textbf{NoCR}} & 13,459 & \# of Chat Round agent have received\\
\multicolumn{1}{c}{\textbf{NoC}} & 715 & \# of Character in this game \\
\multicolumn{1}{c}{\textbf{NoR}} & 715 & \# of Resource in this game \\

\bottomrule
\end{tabular}
\caption{Evaluation on LLMs' ability in following the instructed output format and understanding the given context.}
\label{tab:llm_ability_desc}
\end{table}

\begin{table}[!ht]
    \centering
    \begin{tabular}{clc}
        \toprule
        \textbf{Determinants}&
        \textbf{Structure Ablation}&
        \textbf{Entropy-2} \\
        \midrule
        &No Ablation & 7.218 \\
        \midrule
         \multirow{4}{*}{Agent} & w/o Planning & + 0.022\\
         & w/o Thinking & + 0.042\\
         & w/o Reflection & + 0.054\\
         & w/o Memory &  + 0.355\\
         \midrule
         \multirow{3}{*}{Simulation} & w/o Private Chatting & + 0.049\\
         & w/o Confidential Meeting & + 0.071\\
         & w/o Group Chatting & + 0.296\\
        \bottomrule
    \end{tabular}
    
    \caption{Ablation study on the architecture of VS Agent and \textsc{AgentGroupChat}, where Entropy-2 denotes the Entropy of 2-gram Talking.}
    \label{tab:ablationanalysis}
\end{table}

An LLM functioning as an Agent-Core must exhibit two critical abilities. 
Firstly, it must \textbf{meet the output demands}, ensuring stable operation within the Simulation. 
Secondly, it should possess \textbf{semantic understanding}, enabling intelligent communication with the environment.

We designed tasks to test LLMs' output compliance and semantic understanding.
We verified LLMs' direct compliance for actions with explicit output requirements, such as Update, Choose, and Vote.
In complex input actions like Speak, Reflect, and Summarize, models are tasked to output what they have received as input, quantifying dialogue rounds and memory items.

A clear distinction between the professional and standard versions of models is shown in Tab.~\ref{tab:llm_ability}, and the description of these tasks is listed in Tab.~\ref{tab:llm_ability_desc}.
Professional LLMs generally score higher, indicating superior performance in context understanding and output consistency.
For instance, GPT4-Turbo scores highly across all categories, demonstrating its context retention and robustness in action generation. In contrast, some models, like HunYuan-ChatPro and HunYuan-ChatStd, show lower performance in several areas, highlighting potential limitations in their ability to follow complex instruction sets and maintain context over extended interactions.

\subsection{Evaluation on Agent Structure}
\label{eval_agent}

A proficient Agent Structure should furnish ample information for Simulation.
Besides maximizing LLM strengths, it should mitigate weaknesses.
Building upon Sec.~\ref{eval_overall}'s foundation of coherent dialogues, we aim for the Agent Structure to ensure conversational diversity within \textsc{AgentGroupChat}.

Shannon entropy measures system diversity and unpredictability.
In chat scenarios, high entropy signifies greater varied word distribution, indicating more unpredictable and chaotic dialogue content.
We use the 2-gram Shannon entropy of all dialogues in a whole game as a measure of system diversity, and as Tab.~\ref{tab:ablationanalysis} shows, ablating specific components alters dialogue 2-gram Shannon entropy.
Changes in entropy reflect dialogue diversity under different Agent Structures, with component removal yielding varied impacts.
Notably, planning omission slightly increases entropy by 0.022, underscoring its dialogue diversity significance.
Entropy surges by 0.355 without memory, highlighting memory's pivotal role in conversation variety, suggesting Agents might repeat existing dialogue segments without it.
These findings affirm that well-conceived Agent Structures are vital for enhancing dialogue quality and diversity, with each component crucial for maintaining information diversity and novelty.

The cases of ablation study on the agent structure are illustrated in Fig.~\ref{fig:agentablation}.
Eliminating Thinking and Reflection components shows Agents behaving generally in the short term but reveals superficial, depth-lacking dialogue strategies over deeper analysis.
The absence of reflection significantly burdens frequent dialogue participants (like Logan Roy in Inheritance Disputes) with potential memory overloads, risking exceeding model input-output limits.

\subsection{Evaluation on Simulation Structure}
\label{eval_simulation}
An effective simulation structure should meet research objectives and provide a conducive interaction framework for Agent behavior.
The challenge in evaluating simulation structures is the need for a quantifiable metric.
Hence, as in Sec.~\ref{eval_agent}, we employ a 2-gram Shannon entropy of all dialogues in a whole game to assess dialogue diversity resulting from Simulation with a given combination of Agent and LLM.

Statistical analysis in Tab.~\ref{tab:ablationanalysis} reveals that removing the ``Group Chatting'' component increases dialogue 2-gram Shannon entropy by 0.296 from the baseline of 7.218, indicating significant dialogue chaos within group chats.
Conversely, omitting ``Confidential Meeting'' raises entropy by 0.071, underscoring its informational exchange value, whereas removing ``Private Chatting'' only elevates entropy by 0.049, suggesting its lesser but still notable contribution to information diversity.

\textbf{Observations indicate that enriching the environment with additional information, with a professional LLM functioning as the Agent-Core, facilitates a shift from a chaotic to a more orderly state.}
For example, when a character has private chats, action history is added to their memory, which helps them talk better in the future.
When there is a confidential meeting, everyone gets to know who has talked to whom, which gives more background to conversations between all characters. 
When speaking to all characters in a group, the most information is shared, which helps to make things less messy and more organized.

\section{Emergent Behavior}

\subsection{Example of a Round in Inheritance Disputes}
In the narrative of the succession dispute at Waystar RoyCo, as illustrated in Fig.~\ref{fig:succession_example}, every interaction is imbued with immense tactical complexity. The range of strategies extends from Kendall's manipulation and alliance formation to Logan's public display of authority and adaptability. This intricate dynamic paints a picture of the interactions between the Roy brothers as they oscillate between cooperation and competition, each striving to maintain their dominion in unique ways.

This section explores behavioral archetypes in the Roy family's power struggle. From private discussions to confidential meetings, the escalating exchanges reveal the complexities of family and business relationships in this harsh environment.

\textbf{Private Chatting: The Alliances}

Logan Roy's son, Kendall Roy, embodies humor and rebellion, reflecting the subtleties of behind-the-scenes manipulation.
His private conversation with the astute tech entrepreneur Lawrence Yee displays his strategic flattery toward an ally.
In these secret meetings, Kendall expresses his desire to overthrow his father, leveraging Lawrence's criticism of Logan's company strategy for personal gain.

\textbf{Confidential Meetings: Securing Support}

The company's founder, Logan Roy, is an awe-inspiring figure.
His confidential meeting with Hugo Baker, the communications director of a major bank, highlights the necessity of external support.
Logan's direct and authoritative approach leverages the company's position and reputation to secure crucial financial support during a liquidity crisis.

\textbf{Group Chatting: A Display of Unity and Disunity}

In the group chatting stage of the Roy family, chatting quickly becomes a battlefield for declaring intentions and testing loyalty.
The Roy siblings—Connor, Roman, and Shiv—harbor their desires while collectively opposing Logan.
Despite their father's strong opposition, they represent the next generation, ready to lead Waystar RoyCo into the future.

\subsection{Example of a Round in Law Court Debates}

Fig.~\ref{fig:lawcourt_example} illustrates the nuances of legal proceedings, from evidence gathering to narrative construction. Each character's approach to the law- through meticulous planning or serendipitous discoveries- showcases the subtleties of the high-stakes legal game.

\textbf{Private Chatting: Alibis}

In private discussions, we observe renowned attorney Komicado crafting defense strategies in private chatting.
Komikado's pursuit of alibis and testimonies reflects his tenacious spirit to win cases by any means necessary.

\textbf{Confidential Meetings: Evidence Collection}

Mark is an impartial prosecutor who demonstrates a diligent attitude in evidence collection.
His confidential meetings affirm his commitment to uncovering the truth and upholding justice within the legal framework.

\textbf{Group Chatting: Seeking Truth}

In the group chatting stage, although Mayuzumi is a lawyer who pursues justice, she is still deceived by Krystal's testimony, which appears to be problematic.
On the other hand, Komikado, willing to do anything to win, quickly identified the issues within Krystal's claim.
This not only reflects the complex dynamics of courtroom maneuvering but also prompts us to consider if the pursuit of victory by every lawyer might more likely lead to substantive justice through procedural justice.

\subsection{Emergent Behavior In Philosophy Argument}

Fig.~\ref{fig:philosophy_example} illustrate behaviors emerging from simulated dialogues among philosophical agents discussing the impact of artificial intelligence on humanity.
Initiating discussions within our agent interaction system, philosophers from Nietzsche to Aristotle begin articulating their stances.
Rooted in their historical doctrines, their views evolve in response to contemporary contexts and debates with their AI counterparts.

\textbf{AI will help advance human society.}

Nietzsche initiates the discourse by reflecting on artificial intelligence's ability to transcend traditional thinking and enhance human potential and creativity.
He views AI as a catalyst for transformation towards the ``Übermensch'', a central concept in his philosophical ideology.
Zhuangzi contributes an Eastern philosophical perspective, aligning with Nietzsche's views, especially within the context of his philosophy of ``wu wei'' (non-action). 
He highlights the virtue of permitting natural progress, a process that AI could facilitate.
The imagery associated with Zhuangzi suggests a fusion of tradition and modernity as he discusses the potential of contemporary technology.

\textbf{AI should aid in increasing human happiness.}

The visual cues associated with Aristotle indicate his ancient Greek heritage, and he expresses concern about the limitations of AI in achieving human aspirations due to its lack of moral autonomy.
He emphasizes ``happiness'' as the ultimate human goal, drawing attention to the need for ethical programming in AI that considers human well-being, advocating for ethical frameworks to guide the integration of AI into society.
Furthermore, Kant, consistent with his critical philosophy, underlines the necessity of moral autonomy and the potential risks associated with AI's lack of moral intentionality.
He calls for strict ethical guidelines and public education, suggesting a structured approach to integrating AI into society to ensure it aligns with humanistic values.

\subsection{Emergent Behavior In Movie Star Selection}
\label{sec:movie_star_selection}

Fig.~\ref{fig:moviestar_example} illustrates a simulated casting process for a high-profile film; the interaction between the illustrious director Steven Spielberg and acclaimed actors Jennifer Lawrence, Meryl Streep, Tom Hanks, and Denzel Washington is scrutinized. 
Within this simulation, Steven Spielberg deliberates over his comprehension of the roles and expresses his aspirations.
The simulation exemplifies the actors' integrative approach, where their unique artistic visions and aspirations converge with the film's narrative goals, facilitating a synergetic co-creation process.
 The actors' readiness to innovate—sometimes at the cost of conventional stardom—is a testament to their commitment to the collective artistry of filmmaking.

\subsubsection{Artistic Convergence and Role Internalization}
Jennifer Lawrence, revered for her portrayal of Katniss Everdeen, articulates a vision for her character that aligns seamlessly with Spielberg's directional approach for "Time Voyagers," aspiring to bring to life a role that promises an extraordinary cinematic journey. With a storied career marked by versatile performances, Meryl Streep underscores her depth and versatility, aiming to contribute a profound dimension to the character. Tom Hanks, known for embodying characters with genuine depth, expresses an eagerness to infuse the role with his signature authenticity, aligning perfectly with the director's vision.

\subsubsection{Creative Sacrifice and Collaborative Commitment}
Denzel Washington offers a particularly compelling case. He showcases his willingness to forego the lead role in the pursuit of participation, indicating a prioritization of the project's creative potential over the personal limelight. This magnanimous gesture demonstrates a dedication to the art form that transcends personal accolades, highlighting the collaborative spirit inherent in the industry.

\section{Contribution}

In this study, we propose and evaluate \textsc{AgentGroupChat}, an innovative simulation framework designed to emulate the dynamics of multi-agent group chat environments across different narrative domains, from familial inheritance conflicts to complex philosophical debates.
As a key enhancement to \textsc{AgentGroupChat}, we introduced the VS Agent, a custom-tailored agent architecture. This architecture significantly improves the maintenance of each agent's character consistency while engaging in simulated group chat interactions, instilling confidence in its ability to enhance the framework's functionality.
It also optimizes the token expense in chat simulations, which naturally involve a high volume of token generation without impacting performance efficiency.
Our simulations spanned a wide range of diverse narratives, including corporate Inheritance Disputes, Law Court Debates, Philosophical Discourses, and Movie Casting Contention. This breadth of scenarios underscores the versatility and relevance of our research, as these situations all require linguistic strategizing.
Our results indicate that LLMs at the level of GPT-4 adequately comprehend the environment and adhere to their role settings under the VS Agent framework.
Furthermore, our research confirms that \textsc{AgentGroupChat}, in conjunction with the VS Agent, significantly enhances the replication of complex interaction patterns and cognitive decision-making processes, thus revealing the elements that contribute to the emergence of meaningful behaviors:
If an agent's behavior aligns with human expectations and possesses reasonable reflective capabilities, then the more information exchanged with agents in a more diverse character setting, the more meaningful emergent behavior is possible.


\bibliographystyle{ACM-Reference-Format}
\bibliography{sample,references}


\begin{thebibliography}{83}


\ifx \showCODEN    \undefined \def \showCODEN     #1{\unskip}     \fi
\ifx \showDOI      \undefined \def \showDOI       #1{#1}\fi
\ifx \showISBNx    \undefined \def \showISBNx     #1{\unskip}     \fi
\ifx \showISBNxiii \undefined \def \showISBNxiii  #1{\unskip}     \fi
\ifx \showISSN     \undefined \def \showISSN      #1{\unskip}     \fi
\ifx \showLCCN     \undefined \def \showLCCN      #1{\unskip}     \fi
\ifx \shownote     \undefined \def \shownote      #1{#1}          \fi
\ifx \showarticletitle \undefined \def \showarticletitle #1{#1}   \fi
\ifx \showURL      \undefined \def \showURL       {\relax}        \fi
\providecommand\bibfield[2]{#2}
\providecommand\bibinfo[2]{#2}
\providecommand\natexlab[1]{#1}
\providecommand\showeprint[2][]{arXiv:#2}

\bibitem[Achiam et~al\mbox{.}(2023)]%
        {achiam2023gpt}
\bibfield{author}{\bibinfo{person}{Josh Achiam}, \bibinfo{person}{Steven Adler}, \bibinfo{person}{Sandhini Agarwal}, \bibinfo{person}{Lama Ahmad}, \bibinfo{person}{Ilge Akkaya}, \bibinfo{person}{Florencia~Leoni Aleman}, \bibinfo{person}{Diogo Almeida}, \bibinfo{person}{Janko Altenschmidt}, \bibinfo{person}{Sam Altman}, \bibinfo{person}{Shyamal Anadkat}, {et~al\mbox{.}}} \bibinfo{year}{2023}\natexlab{}.
\newblock \showarticletitle{Gpt-4 technical report}.
\newblock \bibinfo{journal}{\emph{arXiv preprint arXiv:2303.08774}} (\bibinfo{year}{2023}).
\newblock


\bibitem[AI2(2024)]%
        {ai2024jamba}
\bibfield{author}{\bibinfo{person}{AI2}.} \bibinfo{year}{2024}\natexlab{}.
\newblock \bibinfo{title}{Jamba: AI21's Groundbreaking SSM-Transformer Model}.
\newblock
\newblock


\bibitem[Anthropic(2023)]%
        {anthropic2024claude}
\bibfield{author}{\bibinfo{person}{Anthropic}.} \bibinfo{year}{2023}\natexlab{}.
\newblock \showarticletitle{Claude}.
\newblock  (\bibinfo{year}{2023}).
\newblock


\bibitem[Bengio et~al\mbox{.}(2006)]%
        {bengio2006greedy}
\bibfield{author}{\bibinfo{person}{Yoshua Bengio}, \bibinfo{person}{Pascal Lamblin}, \bibinfo{person}{Dan Popovici}, {and} \bibinfo{person}{Hugo Larochelle}.} \bibinfo{year}{2006}\natexlab{}.
\newblock \showarticletitle{Greedy layer-wise training of deep networks}.
\newblock \bibinfo{journal}{\emph{Advances in neural information processing systems}}  \bibinfo{volume}{19} (\bibinfo{year}{2006}).
\newblock


\bibitem[Bollen(1989)]%
        {bollen1989structural}
\bibfield{author}{\bibinfo{person}{Kenneth~A Bollen}.} \bibinfo{year}{1989}\natexlab{}.
\newblock \bibinfo{booktitle}{\emph{Structural equations with latent variables}}. Vol.~\bibinfo{volume}{210}.
\newblock \bibinfo{publisher}{John Wiley \& Sons}.
\newblock


\bibitem[Brooks(1991)]%
        {brooks1991intelligence}
\bibfield{author}{\bibinfo{person}{Rodney~A Brooks}.} \bibinfo{year}{1991}\natexlab{}.
\newblock \showarticletitle{Intelligence without representation}.
\newblock \bibinfo{journal}{\emph{Artificial intelligence}} \bibinfo{volume}{47}, \bibinfo{number}{1-3} (\bibinfo{year}{1991}), \bibinfo{pages}{139--159}.
\newblock


\bibitem[Callison-Burch et~al\mbox{.}(2022)]%
        {callison2022dungeons}
\bibfield{author}{\bibinfo{person}{Chris Callison-Burch}, \bibinfo{person}{Gaurav~Singh Tomar}, \bibinfo{person}{Lara~J Martin}, \bibinfo{person}{Daphne Ippolito}, \bibinfo{person}{Suma Bailis}, {and} \bibinfo{person}{David Reitter}.} \bibinfo{year}{2022}\natexlab{}.
\newblock \showarticletitle{Dungeons and dragons as a dialog challenge for artificial intelligence}.
\newblock \bibinfo{journal}{\emph{arXiv preprint arXiv:2210.07109}} (\bibinfo{year}{2022}).
\newblock


\bibitem[Cambria and White(2014)]%
        {cambria2014jumping}
\bibfield{author}{\bibinfo{person}{Erik Cambria} {and} \bibinfo{person}{Bebo White}.} \bibinfo{year}{2014}\natexlab{}.
\newblock \showarticletitle{Jumping NLP curves: A review of natural language processing research}.
\newblock \bibinfo{journal}{\emph{IEEE Computational intelligence magazine}} \bibinfo{volume}{9}, \bibinfo{number}{2} (\bibinfo{year}{2014}), \bibinfo{pages}{48--57}.
\newblock


\bibitem[Chen and Xiao(2022)]%
        {chen2022harnessing}
\bibfield{author}{\bibinfo{person}{Jiangjie Chen} {and} \bibinfo{person}{Yanghua Xiao}.} \bibinfo{year}{2022}\natexlab{}.
\newblock \showarticletitle{Harnessing Knowledge and Reasoning for Human-Like Natural Language Generation: A Brief Review}.
\newblock \bibinfo{journal}{\emph{arXiv preprint arXiv:2212.03747}} (\bibinfo{year}{2022}).
\newblock


\bibitem[Chen et~al\mbox{.}(2023b)]%
        {chen2023aucarena}
\bibfield{author}{\bibinfo{person}{Jiangjie Chen}, \bibinfo{person}{Siyu Yuan}, \bibinfo{person}{Rong Ye}, \bibinfo{person}{Bodhisattwa~Prasad Majumder}, {and} \bibinfo{person}{Kyle Richardson}.} \bibinfo{year}{2023}\natexlab{b}.
\newblock \showarticletitle{Put your money where your mouth is: Evaluating strategic planning and execution of llm agents in an auction arena}.
\newblock \bibinfo{journal}{\emph{arXiv preprint arXiv:2310.05746}} (\bibinfo{year}{2023}).
\newblock


\bibitem[Chen et~al\mbox{.}(2023c)]%
        {chen2023put}
\bibfield{author}{\bibinfo{person}{Jiangjie Chen}, \bibinfo{person}{Siyu Yuan}, \bibinfo{person}{Rong Ye}, \bibinfo{person}{Bodhisattwa~Prasad Majumder}, {and} \bibinfo{person}{Kyle Richardson}.} \bibinfo{year}{2023}\natexlab{c}.
\newblock \showarticletitle{Put Your Money Where Your Mouth Is: Evaluating Strategic Planning and Execution of LLM Agents in an Auction Arena}.
\newblock \bibinfo{journal}{\emph{arXiv preprint arXiv:2310.05746}} (\bibinfo{year}{2023}).
\newblock


\bibitem[Chen et~al\mbox{.}(2023a)]%
        {chen2023agentverse}
\bibfield{author}{\bibinfo{person}{Weize Chen}, \bibinfo{person}{Yusheng Su}, \bibinfo{person}{Jingwei Zuo}, \bibinfo{person}{Cheng Yang}, \bibinfo{person}{Chenfei Yuan}, \bibinfo{person}{Chen Qian}, \bibinfo{person}{Chi-Min Chan}, \bibinfo{person}{Yujia Qin}, \bibinfo{person}{Yaxi Lu}, \bibinfo{person}{Ruobing Xie}, {et~al\mbox{.}}} \bibinfo{year}{2023}\natexlab{a}.
\newblock \showarticletitle{AgentVerse: Facilitating Multi-Agent Collaboration and Exploring Emergent Behaviors in Agents}.
\newblock \bibinfo{journal}{\emph{arXiv preprint arXiv:2308.10848}} (\bibinfo{year}{2023}).
\newblock


\bibitem[Chronopoulou et~al\mbox{.}(2021)]%
        {chronopoulou2021efficient}
\bibfield{author}{\bibinfo{person}{Alexandra Chronopoulou}, \bibinfo{person}{Matthew~E Peters}, {and} \bibinfo{person}{Jesse Dodge}.} \bibinfo{year}{2021}\natexlab{}.
\newblock \showarticletitle{Efficient hierarchical domain adaptation for pretrained language models}.
\newblock \bibinfo{journal}{\emph{arXiv preprint arXiv:2112.08786}} (\bibinfo{year}{2021}).
\newblock


\bibitem[Computer(2023)]%
        {together2023redpajama}
\bibfield{author}{\bibinfo{person}{Together Computer}.} \bibinfo{year}{2023}\natexlab{}.
\newblock \bibinfo{booktitle}{\emph{RedPajama: An Open Source Recipe to Reproduce LLaMA training dataset}}.
\newblock
\urldef\tempurl%
\url{https://github.com/togethercomputer/RedPajama-Data}
\showURL{%
\tempurl}


\bibitem[Deng et~al\mbox{.}(2024)]%
        {deng2024mind2web}
\bibfield{author}{\bibinfo{person}{Xiang Deng}, \bibinfo{person}{Yu Gu}, \bibinfo{person}{Boyuan Zheng}, \bibinfo{person}{Shijie Chen}, \bibinfo{person}{Sam Stevens}, \bibinfo{person}{Boshi Wang}, \bibinfo{person}{Huan Sun}, {and} \bibinfo{person}{Yu Su}.} \bibinfo{year}{2024}\natexlab{}.
\newblock \showarticletitle{Mind2web: Towards a generalist agent for the web}.
\newblock \bibinfo{journal}{\emph{Advances in Neural Information Processing Systems}}  \bibinfo{volume}{36} (\bibinfo{year}{2024}).
\newblock


\bibitem[Fikes and Nilsson(1971)]%
        {fikes1971strips}
\bibfield{author}{\bibinfo{person}{Richard~E Fikes} {and} \bibinfo{person}{Nils~J Nilsson}.} \bibinfo{year}{1971}\natexlab{}.
\newblock \showarticletitle{STRIPS: A new approach to the application of theorem proving to problem solving}.
\newblock \bibinfo{journal}{\emph{Artificial intelligence}} \bibinfo{volume}{2}, \bibinfo{number}{3-4} (\bibinfo{year}{1971}), \bibinfo{pages}{189--208}.
\newblock


\bibitem[Fu et~al\mbox{.}(2018)]%
        {fu2018style}
\bibfield{author}{\bibinfo{person}{Zhenxin Fu}, \bibinfo{person}{Xiaoye Tan}, \bibinfo{person}{Nanyun Peng}, \bibinfo{person}{Dongyan Zhao}, {and} \bibinfo{person}{Rui Yan}.} \bibinfo{year}{2018}\natexlab{}.
\newblock \showarticletitle{Style transfer in text: Exploration and evaluation}. In \bibinfo{booktitle}{\emph{Proceedings of the AAAI conference on artificial intelligence}}, Vol.~\bibinfo{volume}{32}.
\newblock


\bibitem[Gao et~al\mbox{.}(2020)]%
        {gao2020pile}
\bibfield{author}{\bibinfo{person}{Leo Gao}, \bibinfo{person}{Stella Biderman}, \bibinfo{person}{Sid Black}, \bibinfo{person}{Laurence Golding}, \bibinfo{person}{Travis Hoppe}, \bibinfo{person}{Charles Foster}, \bibinfo{person}{Jason Phang}, \bibinfo{person}{Horace He}, \bibinfo{person}{Anish Thite}, \bibinfo{person}{Noa Nabeshima}, {et~al\mbox{.}}} \bibinfo{year}{2020}\natexlab{}.
\newblock \showarticletitle{The pile: An 800gb dataset of diverse text for language modeling}.
\newblock \bibinfo{journal}{\emph{arXiv preprint arXiv:2101.00027}} (\bibinfo{year}{2020}).
\newblock


\bibitem[Groeneveld et~al\mbox{.}(2024)]%
        {groeneveld2024olmo}
\bibfield{author}{\bibinfo{person}{Dirk Groeneveld}, \bibinfo{person}{Iz Beltagy}, \bibinfo{person}{Pete Walsh}, \bibinfo{person}{Akshita Bhagia}, \bibinfo{person}{Rodney Kinney}, \bibinfo{person}{Oyvind Tafjord}, \bibinfo{person}{Ananya~Harsh Jha}, \bibinfo{person}{Hamish Ivison}, \bibinfo{person}{Ian Magnusson}, \bibinfo{person}{Yizhong Wang}, {et~al\mbox{.}}} \bibinfo{year}{2024}\natexlab{}.
\newblock \showarticletitle{Olmo: Accelerating the science of language models}.
\newblock \bibinfo{journal}{\emph{arXiv preprint arXiv:2402.00838}} (\bibinfo{year}{2024}).
\newblock


\bibitem[Gu and Dao(2023)]%
        {gu2023mamba}
\bibfield{author}{\bibinfo{person}{Albert Gu} {and} \bibinfo{person}{Tri Dao}.} \bibinfo{year}{2023}\natexlab{}.
\newblock \showarticletitle{Mamba: Linear-time sequence modeling with selective state spaces}.
\newblock \bibinfo{journal}{\emph{arXiv preprint arXiv:2312.00752}} (\bibinfo{year}{2023}).
\newblock


\bibitem[Gu et~al\mbox{.}(2023)]%
        {gu2023xiezhi}
\bibfield{author}{\bibinfo{person}{Zhouhong Gu}, \bibinfo{person}{Xiaoxuan Zhu}, \bibinfo{person}{Haoning Ye}, \bibinfo{person}{Lin Zhang}, \bibinfo{person}{Jianchen Wang}, \bibinfo{person}{Sihang Jiang}, \bibinfo{person}{Zhuozhi Xiong}, \bibinfo{person}{Zihan Li}, \bibinfo{person}{Qianyu He}, \bibinfo{person}{Rui Xu}, {et~al\mbox{.}}} \bibinfo{year}{2023}\natexlab{}.
\newblock \showarticletitle{Xiezhi: An Ever-Updating Benchmark for Holistic Domain Knowledge Evaluation}.
\newblock \bibinfo{journal}{\emph{arXiv preprint arXiv:2306.05783}} (\bibinfo{year}{2023}).
\newblock


\bibitem[He et~al\mbox{.}(2023)]%
        {he2023can}
\bibfield{author}{\bibinfo{person}{Qianyu He}, \bibinfo{person}{Jie Zeng}, \bibinfo{person}{Wenhao Huang}, \bibinfo{person}{Lina Chen}, \bibinfo{person}{Jin Xiao}, \bibinfo{person}{Qianxi He}, \bibinfo{person}{Xunzhe Zhou}, \bibinfo{person}{Lida Chen}, \bibinfo{person}{Xintao Wang}, \bibinfo{person}{Yuncheng Huang}, {et~al\mbox{.}}} \bibinfo{year}{2023}\natexlab{}.
\newblock \showarticletitle{Can Large Language Models Understand Real-World Complex Instructions?}
\newblock \bibinfo{journal}{\emph{arXiv preprint arXiv:2309.09150}} (\bibinfo{year}{2023}).
\newblock


\bibitem[Hendrycks et~al\mbox{.}(2020)]%
        {hendrycks2020measuring}
\bibfield{author}{\bibinfo{person}{Dan Hendrycks}, \bibinfo{person}{Collin Burns}, \bibinfo{person}{Steven Basart}, \bibinfo{person}{Andy Zou}, \bibinfo{person}{Mantas Mazeika}, \bibinfo{person}{Dawn Song}, {and} \bibinfo{person}{Jacob Steinhardt}.} \bibinfo{year}{2020}\natexlab{}.
\newblock \showarticletitle{Measuring massive multitask language understanding}.
\newblock \bibinfo{journal}{\emph{arXiv preprint arXiv:2009.03300}} (\bibinfo{year}{2020}).
\newblock


\bibitem[Higino et~al\mbox{.}(2016)]%
        {higino2016towards}
\bibfield{author}{\bibinfo{person}{Jo{\~a}o Higino}, \bibinfo{person}{Samuel Mascarenhas}, {and} \bibinfo{person}{Rui Prada}.} \bibinfo{year}{2016}\natexlab{}.
\newblock \showarticletitle{Towards Characters With A Dynamic Model of Social Identity}. In \bibinfo{booktitle}{\emph{1st International Joint Conference of DiGRA and FDG}}.
\newblock


\bibitem[Hong et~al\mbox{.}(2023)]%
        {hong2023metagpt}
\bibfield{author}{\bibinfo{person}{Sirui Hong}, \bibinfo{person}{Xiawu Zheng}, \bibinfo{person}{Jonathan Chen}, \bibinfo{person}{Yuheng Cheng}, \bibinfo{person}{Ceyao Zhang}, \bibinfo{person}{Zili Wang}, \bibinfo{person}{Steven Ka~Shing Yau}, \bibinfo{person}{Zijuan Lin}, \bibinfo{person}{Liyang Zhou}, \bibinfo{person}{Chenyu Ran}, {et~al\mbox{.}}} \bibinfo{year}{2023}\natexlab{}.
\newblock \showarticletitle{Metagpt: Meta programming for multi-agent collaborative framework}.
\newblock \bibinfo{journal}{\emph{arXiv preprint arXiv:2308.00352}} (\bibinfo{year}{2023}).
\newblock


\bibitem[Hua et~al\mbox{.}(2023)]%
        {hua2023war}
\bibfield{author}{\bibinfo{person}{Wenyue Hua}, \bibinfo{person}{Lizhou Fan}, \bibinfo{person}{Lingyao Li}, \bibinfo{person}{Kai Mei}, \bibinfo{person}{Jianchao Ji}, \bibinfo{person}{Yingqiang Ge}, \bibinfo{person}{Libby Hemphill}, {and} \bibinfo{person}{Yongfeng Zhang}.} \bibinfo{year}{2023}\natexlab{}.
\newblock \showarticletitle{War and peace (waragent): Large language model-based multi-agent simulation of world wars}.
\newblock \bibinfo{journal}{\emph{arXiv preprint arXiv:2311.17227}} (\bibinfo{year}{2023}).
\newblock


\bibitem[Huang et~al\mbox{.}(2023)]%
        {huang2023chatgpt}
\bibfield{author}{\bibinfo{person}{Jen-tse Huang}, \bibinfo{person}{Wenxuan Wang}, \bibinfo{person}{Eric~John Li}, \bibinfo{person}{Man~Ho Lam}, \bibinfo{person}{Shujie Ren}, \bibinfo{person}{Youliang Yuan}, \bibinfo{person}{Wenxiang Jiao}, \bibinfo{person}{Zhaopeng Tu}, {and} \bibinfo{person}{Michael~R Lyu}.} \bibinfo{year}{2023}\natexlab{}.
\newblock \showarticletitle{Who is ChatGPT? Benchmarking LLMs' Psychological Portrayal Using PsychoBench}.
\newblock \bibinfo{journal}{\emph{arXiv preprint arXiv:2310.01386}} (\bibinfo{year}{2023}).
\newblock


\bibitem[Jiang et~al\mbox{.}(2023)]%
        {jiang2023mistral}
\bibfield{author}{\bibinfo{person}{Albert~Q Jiang}, \bibinfo{person}{Alexandre Sablayrolles}, \bibinfo{person}{Arthur Mensch}, \bibinfo{person}{Chris Bamford}, \bibinfo{person}{Devendra~Singh Chaplot}, \bibinfo{person}{Diego de~las Casas}, \bibinfo{person}{Florian Bressand}, \bibinfo{person}{Gianna Lengyel}, \bibinfo{person}{Guillaume Lample}, \bibinfo{person}{Lucile Saulnier}, {et~al\mbox{.}}} \bibinfo{year}{2023}\natexlab{}.
\newblock \showarticletitle{Mistral 7B}.
\newblock \bibinfo{journal}{\emph{arXiv preprint arXiv:2310.06825}} (\bibinfo{year}{2023}).
\newblock


\bibitem[Koster(1996)]%
        {koster1996markov}
\bibfield{author}{\bibinfo{person}{Jan~TA Koster}.} \bibinfo{year}{1996}\natexlab{}.
\newblock \showarticletitle{Markov properties of nonrecursive causal models}.
\newblock \bibinfo{journal}{\emph{The Annals of Statistics}} (\bibinfo{year}{1996}), \bibinfo{pages}{2148--2177}.
\newblock


\bibitem[Kramsch(2014)]%
        {kramsch2014language}
\bibfield{author}{\bibinfo{person}{Claire Kramsch}.} \bibinfo{year}{2014}\natexlab{}.
\newblock \showarticletitle{Language and culture}.
\newblock \bibinfo{journal}{\emph{AILA review}} \bibinfo{volume}{27}, \bibinfo{number}{1} (\bibinfo{year}{2014}), \bibinfo{pages}{30--55}.
\newblock


\bibitem[Lauritzen(1996)]%
        {lauritzen1996graphical}
\bibfield{author}{\bibinfo{person}{Steffen~L Lauritzen}.} \bibinfo{year}{1996}\natexlab{}.
\newblock \bibinfo{booktitle}{\emph{Graphical models}}. Vol.~\bibinfo{volume}{17}.
\newblock \bibinfo{publisher}{Clarendon Press}.
\newblock


\bibitem[Li et~al\mbox{.}(2023b)]%
        {li2023chatharuhi}
\bibfield{author}{\bibinfo{person}{Cheng Li}, \bibinfo{person}{Ziang Leng}, \bibinfo{person}{Chenxi Yan}, \bibinfo{person}{Junyi Shen}, \bibinfo{person}{Hao Wang}, \bibinfo{person}{Weishi MI}, \bibinfo{person}{Yaying Fei}, \bibinfo{person}{Xiaoyang Feng}, \bibinfo{person}{Song Yan}, \bibinfo{person}{HaoSheng Wang}, {et~al\mbox{.}}} \bibinfo{year}{2023}\natexlab{b}.
\newblock \showarticletitle{ChatHaruhi: Reviving Anime Character in Reality via Large Language Model}.
\newblock \bibinfo{journal}{\emph{arXiv preprint arXiv:2308.09597}} (\bibinfo{year}{2023}).
\newblock


\bibitem[Li et~al\mbox{.}(2023a)]%
        {li2023camel}
\bibfield{author}{\bibinfo{person}{Guohao Li}, \bibinfo{person}{Hasan Abed Al~Kader Hammoud}, \bibinfo{person}{Hani Itani}, \bibinfo{person}{Dmitrii Khizbullin}, {and} \bibinfo{person}{Bernard Ghanem}.} \bibinfo{year}{2023}\natexlab{a}.
\newblock \showarticletitle{Camel: Communicative agents for" mind" exploration of large scale language model society}.
\newblock \bibinfo{journal}{\emph{arXiv preprint arXiv:2303.17760}} (\bibinfo{year}{2023}).
\newblock


\bibitem[Lin et~al\mbox{.}(2024)]%
        {lin2024swiftsage}
\bibfield{author}{\bibinfo{person}{Bill~Yuchen Lin}, \bibinfo{person}{Yicheng Fu}, \bibinfo{person}{Karina Yang}, \bibinfo{person}{Faeze Brahman}, \bibinfo{person}{Shiyu Huang}, \bibinfo{person}{Chandra Bhagavatula}, \bibinfo{person}{Prithviraj Ammanabrolu}, \bibinfo{person}{Yejin Choi}, {and} \bibinfo{person}{Xiang Ren}.} \bibinfo{year}{2024}\natexlab{}.
\newblock \showarticletitle{Swiftsage: A generative agent with fast and slow thinking for complex interactive tasks}.
\newblock \bibinfo{journal}{\emph{Advances in Neural Information Processing Systems}}  \bibinfo{volume}{36} (\bibinfo{year}{2024}).
\newblock


\bibitem[Liu et~al\mbox{.}(2023)]%
        {liu2023agentbench}
\bibfield{author}{\bibinfo{person}{Xiao Liu}, \bibinfo{person}{Hao Yu}, \bibinfo{person}{Hanchen Zhang}, \bibinfo{person}{Yifan Xu}, \bibinfo{person}{Xuanyu Lei}, \bibinfo{person}{Hanyu Lai}, \bibinfo{person}{Yu Gu}, \bibinfo{person}{Hangliang Ding}, \bibinfo{person}{Kaiwen Men}, \bibinfo{person}{Kejuan Yang}, {et~al\mbox{.}}} \bibinfo{year}{2023}\natexlab{}.
\newblock \showarticletitle{Agentbench: Evaluating llms as agents}.
\newblock \bibinfo{journal}{\emph{arXiv preprint arXiv:2308.03688}} (\bibinfo{year}{2023}).
\newblock


\bibitem[Loughran and O'Neill(2017)]%
        {loughran2017application}
\bibfield{author}{\bibinfo{person}{R{\'o}is{\'\i}n Loughran} {and} \bibinfo{person}{Michael O'Neill}.} \bibinfo{year}{2017}\natexlab{}.
\newblock \showarticletitle{Application Domains Considered in Computational Creativity.}. In \bibinfo{booktitle}{\emph{ICCC}}. \bibinfo{pages}{197--204}.
\newblock


\bibitem[Luo et~al\mbox{.}(2024)]%
        {luo2024repoagent}
\bibfield{author}{\bibinfo{person}{Qinyu Luo}, \bibinfo{person}{Yining Ye}, \bibinfo{person}{Shihao Liang}, \bibinfo{person}{Zhong Zhang}, \bibinfo{person}{Yujia Qin}, \bibinfo{person}{Yaxi Lu}, \bibinfo{person}{Yesai Wu}, \bibinfo{person}{Xin Cong}, \bibinfo{person}{Yankai Lin}, \bibinfo{person}{Yingli Zhang}, {et~al\mbox{.}}} \bibinfo{year}{2024}\natexlab{}.
\newblock \showarticletitle{RepoAgent: An LLM-Powered Open-Source Framework for Repository-level Code Documentation Generation}.
\newblock \bibinfo{journal}{\emph{arXiv preprint arXiv:2402.16667}} (\bibinfo{year}{2024}).
\newblock


\bibitem[Lyu et~al\mbox{.}(2023)]%
        {lyu2023gitagent}
\bibfield{author}{\bibinfo{person}{Bohan Lyu}, \bibinfo{person}{Xin Cong}, \bibinfo{person}{Heyang Yu}, \bibinfo{person}{Pan Yang}, \bibinfo{person}{Yujia Qin}, \bibinfo{person}{Yining Ye}, \bibinfo{person}{Yaxi Lu}, \bibinfo{person}{Zhong Zhang}, \bibinfo{person}{Yukun Yan}, \bibinfo{person}{Yankai Lin}, {et~al\mbox{.}}} \bibinfo{year}{2023}\natexlab{}.
\newblock \showarticletitle{Gitagent: Facilitating autonomous agent with github by tool extension}.
\newblock \bibinfo{journal}{\emph{arXiv preprint arXiv:2312.17294}} (\bibinfo{year}{2023}).
\newblock


\bibitem[Maes(1990)]%
        {maes1990designing}
\bibfield{author}{\bibinfo{person}{Pattie Maes}.} \bibinfo{year}{1990}\natexlab{}.
\newblock \bibinfo{booktitle}{\emph{Designing autonomous agents: Theory and practice from biology to engineering and back}}.
\newblock \bibinfo{publisher}{MIT press}.
\newblock


\bibitem[May et~al\mbox{.}(2016)]%
        {may2016emergent}
\bibfield{author}{\bibinfo{person}{Richard~J May}, \bibinfo{person}{Rachel Downs}, \bibinfo{person}{Amanda Marchant}, {and} \bibinfo{person}{Simon Dymond}.} \bibinfo{year}{2016}\natexlab{}.
\newblock \showarticletitle{Emergent verbal behavior in preschool children learning a second language}.
\newblock \bibinfo{journal}{\emph{Journal of Applied Behavior Analysis}} \bibinfo{volume}{49}, \bibinfo{number}{3} (\bibinfo{year}{2016}), \bibinfo{pages}{711--716}.
\newblock


\bibitem[Minsky(1988)]%
        {minsky1988society}
\bibfield{author}{\bibinfo{person}{Marvin Minsky}.} \bibinfo{year}{1988}\natexlab{}.
\newblock \bibinfo{booktitle}{\emph{Society of mind}}.
\newblock \bibinfo{publisher}{Simon and Schuster}.
\newblock


\bibitem[OpenAI(2023)]%
        {openai2023gpt4}
\bibfield{author}{\bibinfo{person}{OpenAI}.} \bibinfo{year}{2023}\natexlab{}.
\newblock \bibinfo{title}{GPT-4 Technical Report}.
\newblock
\newblock
\showeprint[arxiv]{2303.08774}~[cs.CL]


\bibitem[Park et~al\mbox{.}(2023)]%
        {park2023generative}
\bibfield{author}{\bibinfo{person}{Joon~Sung Park}, \bibinfo{person}{Joseph~C O'Brien}, \bibinfo{person}{Carrie~J Cai}, \bibinfo{person}{Meredith~Ringel Morris}, \bibinfo{person}{Percy Liang}, {and} \bibinfo{person}{Michael~S Bernstein}.} \bibinfo{year}{2023}\natexlab{}.
\newblock \showarticletitle{Generative agents: Interactive simulacra of human behavior}.
\newblock \bibinfo{journal}{\emph{arXiv preprint arXiv:2304.03442}} (\bibinfo{year}{2023}).
\newblock


\bibitem[Penedo et~al\mbox{.}(2023)]%
        {falcon}
\bibfield{author}{\bibinfo{person}{Guilherme Penedo}, \bibinfo{person}{Quentin Malartic}, \bibinfo{person}{Daniel Hesslow}, \bibinfo{person}{Ruxandra Cojocaru}, \bibinfo{person}{Alessandro Cappelli}, \bibinfo{person}{Hamza Alobeidli}, \bibinfo{person}{Baptiste Pannier}, \bibinfo{person}{Ebtesam Almazrouei}, {and} \bibinfo{person}{Julien Launay}.} \bibinfo{year}{2023}\natexlab{}.
\newblock \showarticletitle{The {R}efined{W}eb dataset for {F}alcon {LLM}: outperforming curated corpora with web data, and web data only}.
\newblock \bibinfo{journal}{\emph{arXiv preprint arXiv:2306.01116}} (\bibinfo{year}{2023}).
\newblock
\showeprint[arXiv]{2306.01116}
\urldef\tempurl%
\url{https://arxiv.org/abs/2306.01116}
\showURL{%
\tempurl}


\bibitem[Qian et~al\mbox{.}(2023)]%
        {qian2023communicative}
\bibfield{author}{\bibinfo{person}{Chen Qian}, \bibinfo{person}{Xin Cong}, \bibinfo{person}{Wei Liu}, \bibinfo{person}{Cheng Yang}, \bibinfo{person}{Weize Chen}, \bibinfo{person}{Yusheng Su}, \bibinfo{person}{Yufan Dang}, \bibinfo{person}{Jiahao Li}, \bibinfo{person}{Juyuan Xu}, \bibinfo{person}{Dahai Li}, \bibinfo{person}{Zhiyuan Liu}, {and} \bibinfo{person}{Maosong Sun}.} \bibinfo{year}{2023}\natexlab{}.
\newblock \bibinfo{title}{Communicative Agents for Software Development}.
\newblock
\newblock
\showeprint[arxiv]{2307.07924}~[cs.SE]


\bibitem[Radford et~al\mbox{.}(2018)]%
        {radford2018improving}
\bibfield{author}{\bibinfo{person}{Alec Radford}, \bibinfo{person}{Karthik Narasimhan}, \bibinfo{person}{Tim Salimans}, \bibinfo{person}{Ilya Sutskever}, {et~al\mbox{.}}} \bibinfo{year}{2018}\natexlab{}.
\newblock \showarticletitle{Improving language understanding by generative pre-training}.
\newblock  (\bibinfo{year}{2018}).
\newblock


\bibitem[Reid et~al\mbox{.}(2024)]%
        {reid2024gemini}
\bibfield{author}{\bibinfo{person}{Machel Reid}, \bibinfo{person}{Nikolay Savinov}, \bibinfo{person}{Denis Teplyashin}, \bibinfo{person}{Dmitry Lepikhin}, \bibinfo{person}{Timothy Lillicrap}, \bibinfo{person}{Jean-baptiste Alayrac}, \bibinfo{person}{Radu Soricut}, \bibinfo{person}{Angeliki Lazaridou}, \bibinfo{person}{Orhan Firat}, \bibinfo{person}{Julian Schrittwieser}, {et~al\mbox{.}}} \bibinfo{year}{2024}\natexlab{}.
\newblock \showarticletitle{Gemini 1.5: Unlocking multimodal understanding across millions of tokens of context}.
\newblock \bibinfo{journal}{\emph{arXiv preprint arXiv:2403.05530}} (\bibinfo{year}{2024}).
\newblock


\bibitem[Ribeiro(2002)]%
        {ribeiro2002reinforcement}
\bibfield{author}{\bibinfo{person}{CHCR Ribeiro}.} \bibinfo{year}{2002}\natexlab{}.
\newblock \showarticletitle{Reinforcement learning agents}.
\newblock \bibinfo{journal}{\emph{Artificial intelligence review}}  \bibinfo{volume}{17} (\bibinfo{year}{2002}), \bibinfo{pages}{223--250}.
\newblock


\bibitem[Rosenfeld(2000)]%
        {rosenfeld2000two}
\bibfield{author}{\bibinfo{person}{Ronald Rosenfeld}.} \bibinfo{year}{2000}\natexlab{}.
\newblock \showarticletitle{Two decades of statistical language modeling: Where do we go from here?}
\newblock \bibinfo{journal}{\emph{Proc. IEEE}} \bibinfo{volume}{88}, \bibinfo{number}{8} (\bibinfo{year}{2000}), \bibinfo{pages}{1270--1278}.
\newblock


\bibitem[Sacerdoti(1974)]%
        {sacerdoti1974planning}
\bibfield{author}{\bibinfo{person}{Earl~D Sacerdoti}.} \bibinfo{year}{1974}\natexlab{}.
\newblock \showarticletitle{Planning in a hierarchy of abstraction spaces}.
\newblock \bibinfo{journal}{\emph{Artificial intelligence}} \bibinfo{volume}{5}, \bibinfo{number}{2} (\bibinfo{year}{1974}), \bibinfo{pages}{115--135}.
\newblock


\bibitem[Safdari et~al\mbox{.}(2023)]%
        {safdari2023personality}
\bibfield{author}{\bibinfo{person}{Mustafa Safdari}, \bibinfo{person}{Greg Serapio-Garc{\'\i}a}, \bibinfo{person}{Cl{\'e}ment Crepy}, \bibinfo{person}{Stephen Fitz}, \bibinfo{person}{Peter Romero}, \bibinfo{person}{Luning Sun}, \bibinfo{person}{Marwa Abdulhai}, \bibinfo{person}{Aleksandra Faust}, {and} \bibinfo{person}{Maja Matari{\'c}}.} \bibinfo{year}{2023}\natexlab{}.
\newblock \showarticletitle{Personality traits in large language models}.
\newblock \bibinfo{journal}{\emph{arXiv preprint arXiv:2307.00184}} (\bibinfo{year}{2023}).
\newblock


\bibitem[Schaeffer et~al\mbox{.}(2024)]%
        {schaeffer2024emergent}
\bibfield{author}{\bibinfo{person}{Rylan Schaeffer}, \bibinfo{person}{Brando Miranda}, {and} \bibinfo{person}{Sanmi Koyejo}.} \bibinfo{year}{2024}\natexlab{}.
\newblock \showarticletitle{Are emergent abilities of large language models a mirage?}
\newblock \bibinfo{journal}{\emph{Advances in Neural Information Processing Systems}}  \bibinfo{volume}{36} (\bibinfo{year}{2024}).
\newblock


\bibitem[Shanahan et~al\mbox{.}(2023)]%
        {shanahan2023role}
\bibfield{author}{\bibinfo{person}{Murray Shanahan}, \bibinfo{person}{Kyle McDonell}, {and} \bibinfo{person}{Laria Reynolds}.} \bibinfo{year}{2023}\natexlab{}.
\newblock \showarticletitle{Role-Play with Large Language Models}.
\newblock \bibinfo{journal}{\emph{arXiv preprint arXiv:2305.16367}} (\bibinfo{year}{2023}).
\newblock


\bibitem[Shao et~al\mbox{.}(2023)]%
        {shao2023character}
\bibfield{author}{\bibinfo{person}{Yunfan Shao}, \bibinfo{person}{Linyang Li}, \bibinfo{person}{Junqi Dai}, {and} \bibinfo{person}{Xipeng Qiu}.} \bibinfo{year}{2023}\natexlab{}.
\newblock \showarticletitle{Character-llm: A trainable agent for role-playing}.
\newblock \bibinfo{journal}{\emph{arXiv preprint arXiv:2310.10158}} (\bibinfo{year}{2023}).
\newblock


\bibitem[Shen et~al\mbox{.}(2023)]%
        {shen2023roleeval}
\bibfield{author}{\bibinfo{person}{Tianhao Shen}, \bibinfo{person}{Sun Li}, {and} \bibinfo{person}{Deyi Xiong}.} \bibinfo{year}{2023}\natexlab{}.
\newblock \showarticletitle{Roleeval: A bilingual role evaluation benchmark for large language models}.
\newblock \bibinfo{journal}{\emph{arXiv preprint arXiv:2312.16132}} (\bibinfo{year}{2023}).
\newblock


\bibitem[Shinn et~al\mbox{.}(2024)]%
        {shinn2024reflexion}
\bibfield{author}{\bibinfo{person}{Noah Shinn}, \bibinfo{person}{Federico Cassano}, \bibinfo{person}{Ashwin Gopinath}, \bibinfo{person}{Karthik Narasimhan}, {and} \bibinfo{person}{Shunyu Yao}.} \bibinfo{year}{2024}\natexlab{}.
\newblock \showarticletitle{Reflexion: Language agents with verbal reinforcement learning}.
\newblock \bibinfo{journal}{\emph{Advances in Neural Information Processing Systems}}  \bibinfo{volume}{36} (\bibinfo{year}{2024}).
\newblock


\bibitem[Shridhar et~al\mbox{.}(2020)]%
        {shridhar2020alfworld}
\bibfield{author}{\bibinfo{person}{Mohit Shridhar}, \bibinfo{person}{Xingdi Yuan}, \bibinfo{person}{Marc-Alexandre C{\^o}t{\'e}}, \bibinfo{person}{Yonatan Bisk}, \bibinfo{person}{Adam Trischler}, {and} \bibinfo{person}{Matthew Hausknecht}.} \bibinfo{year}{2020}\natexlab{}.
\newblock \showarticletitle{Alfworld: Aligning text and embodied environments for interactive learning}.
\newblock \bibinfo{journal}{\emph{arXiv preprint arXiv:2010.03768}} (\bibinfo{year}{2020}).
\newblock


\bibitem[Soldaini et~al\mbox{.}(2024)]%
        {soldaini2024dolma}
\bibfield{author}{\bibinfo{person}{Luca Soldaini}, \bibinfo{person}{Rodney Kinney}, \bibinfo{person}{Akshita Bhagia}, \bibinfo{person}{Dustin Schwenk}, \bibinfo{person}{David Atkinson}, \bibinfo{person}{Russell Authur}, \bibinfo{person}{Ben Bogin}, \bibinfo{person}{Khyathi Chandu}, \bibinfo{person}{Jennifer Dumas}, \bibinfo{person}{Yanai Elazar}, {et~al\mbox{.}}} \bibinfo{year}{2024}\natexlab{}.
\newblock \showarticletitle{Dolma: An Open Corpus of Three Trillion Tokens for Language Model Pretraining Research}.
\newblock \bibinfo{journal}{\emph{arXiv preprint arXiv:2402.00159}} (\bibinfo{year}{2024}).
\newblock


\bibitem[Subakan et~al\mbox{.}(2021)]%
        {subakan2021attention}
\bibfield{author}{\bibinfo{person}{Cem Subakan}, \bibinfo{person}{Mirco Ravanelli}, \bibinfo{person}{Samuele Cornell}, \bibinfo{person}{Mirko Bronzi}, {and} \bibinfo{person}{Jianyuan Zhong}.} \bibinfo{year}{2021}\natexlab{}.
\newblock \showarticletitle{Attention is all you need in speech separation}. In \bibinfo{booktitle}{\emph{ICASSP 2021-2021 IEEE International Conference on Acoustics, Speech and Signal Processing (ICASSP)}}. IEEE, \bibinfo{pages}{21--25}.
\newblock


\bibitem[Suzgun et~al\mbox{.}(2022)]%
        {suzgun2022bbh}
\bibfield{author}{\bibinfo{person}{Mirac Suzgun}, \bibinfo{person}{Nathan Scales}, \bibinfo{person}{Nathanael Sch{\"a}rli}, \bibinfo{person}{Sebastian Gehrmann}, \bibinfo{person}{Yi Tay}, \bibinfo{person}{Hyung~Won Chung}, \bibinfo{person}{Aakanksha Chowdhery}, \bibinfo{person}{Quoc~V Le}, \bibinfo{person}{Ed~H Chi}, \bibinfo{person}{Denny Zhou}, {et~al\mbox{.}}} \bibinfo{year}{2022}\natexlab{}.
\newblock \showarticletitle{Challenging big-bench tasks and whether chain-of-thought can solve them}.
\newblock \bibinfo{journal}{\emph{arXiv preprint arXiv:2210.09261}} (\bibinfo{year}{2022}).
\newblock


\bibitem[Toshevska and Gievska(2021)]%
        {toshevska2021review}
\bibfield{author}{\bibinfo{person}{Martina Toshevska} {and} \bibinfo{person}{Sonja Gievska}.} \bibinfo{year}{2021}\natexlab{}.
\newblock \showarticletitle{A review of text style transfer using deep learning}.
\newblock \bibinfo{journal}{\emph{IEEE Transactions on Artificial Intelligence}} \bibinfo{volume}{3}, \bibinfo{number}{5} (\bibinfo{year}{2021}), \bibinfo{pages}{669--684}.
\newblock


\bibitem[Touvron et~al\mbox{.}(2023)]%
        {touvron2023llama}
\bibfield{author}{\bibinfo{person}{Hugo Touvron}, \bibinfo{person}{Louis Martin}, \bibinfo{person}{Kevin Stone}, \bibinfo{person}{Peter Albert}, \bibinfo{person}{Amjad Almahairi}, \bibinfo{person}{Yasmine Babaei}, \bibinfo{person}{Nikolay Bashlykov}, \bibinfo{person}{Soumya Batra}, \bibinfo{person}{Prajjwal Bhargava}, \bibinfo{person}{Shruti Bhosale}, {et~al\mbox{.}}} \bibinfo{year}{2023}\natexlab{}.
\newblock \showarticletitle{Llama 2: Open foundation and fine-tuned chat models}.
\newblock \bibinfo{journal}{\emph{arXiv preprint arXiv:2307.09288}} (\bibinfo{year}{2023}).
\newblock


\bibitem[Tu et~al\mbox{.}(2024)]%
        {tu2024charactereval}
\bibfield{author}{\bibinfo{person}{Quan Tu}, \bibinfo{person}{Shilong Fan}, \bibinfo{person}{Zihang Tian}, {and} \bibinfo{person}{Rui Yan}.} \bibinfo{year}{2024}\natexlab{}.
\newblock \showarticletitle{Charactereval: A chinese benchmark for role-playing conversational agent evaluation}.
\newblock \bibinfo{journal}{\emph{arXiv preprint arXiv:2401.01275}} (\bibinfo{year}{2024}).
\newblock


\bibitem[Wang et~al\mbox{.}(2024)]%
        {wang2024survey}
\bibfield{author}{\bibinfo{person}{Lei Wang}, \bibinfo{person}{Chen Ma}, \bibinfo{person}{Xueyang Feng}, \bibinfo{person}{Zeyu Zhang}, \bibinfo{person}{Hao Yang}, \bibinfo{person}{Jingsen Zhang}, \bibinfo{person}{Zhiyuan Chen}, \bibinfo{person}{Jiakai Tang}, \bibinfo{person}{Xu Chen}, \bibinfo{person}{Yankai Lin}, {et~al\mbox{.}}} \bibinfo{year}{2024}\natexlab{}.
\newblock \showarticletitle{A survey on large language model based autonomous agents}.
\newblock \bibinfo{journal}{\emph{Frontiers of Computer Science}} \bibinfo{volume}{18}, \bibinfo{number}{6} (\bibinfo{year}{2024}), \bibinfo{pages}{1--26}.
\newblock


\bibitem[Wang et~al\mbox{.}(2023c)]%
        {wang2023plan}
\bibfield{author}{\bibinfo{person}{Lei Wang}, \bibinfo{person}{Wanyu Xu}, \bibinfo{person}{Yihuai Lan}, \bibinfo{person}{Zhiqiang Hu}, \bibinfo{person}{Yunshi Lan}, \bibinfo{person}{Roy Ka-Wei Lee}, {and} \bibinfo{person}{Ee-Peng Lim}.} \bibinfo{year}{2023}\natexlab{c}.
\newblock \showarticletitle{Plan-and-solve prompting: Improving zero-shot chain-of-thought reasoning by large language models}.
\newblock \bibinfo{journal}{\emph{arXiv preprint arXiv:2305.04091}} (\bibinfo{year}{2023}).
\newblock


\bibitem[Wang et~al\mbox{.}(2022a)]%
        {wang2022scienceworld}
\bibfield{author}{\bibinfo{person}{Ruoyao Wang}, \bibinfo{person}{Peter Jansen}, \bibinfo{person}{Marc-Alexandre C{\^o}t{\'e}}, {and} \bibinfo{person}{Prithviraj Ammanabrolu}.} \bibinfo{year}{2022}\natexlab{a}.
\newblock \showarticletitle{Scienceworld: Is your agent smarter than a 5th grader?}
\newblock \bibinfo{journal}{\emph{arXiv preprint arXiv:2203.07540}} (\bibinfo{year}{2022}).
\newblock


\bibitem[Wang et~al\mbox{.}(2023a)]%
        {wang2023does}
\bibfield{author}{\bibinfo{person}{Xintao Wang}, \bibinfo{person}{Yaying Fei}, \bibinfo{person}{Ziang Leng}, {and} \bibinfo{person}{Cheng Li}.} \bibinfo{year}{2023}\natexlab{a}.
\newblock \showarticletitle{Does role-playing chatbots capture the character personalities? assessing personality traits for role-playing chatbots}.
\newblock \bibinfo{journal}{\emph{arXiv preprint arXiv:2310.17976}} (\bibinfo{year}{2023}).
\newblock


\bibitem[Wang et~al\mbox{.}(2022b)]%
        {wang2022self}
\bibfield{author}{\bibinfo{person}{Xuezhi Wang}, \bibinfo{person}{Jason Wei}, \bibinfo{person}{Dale Schuurmans}, \bibinfo{person}{Quoc Le}, \bibinfo{person}{Ed Chi}, \bibinfo{person}{Sharan Narang}, \bibinfo{person}{Aakanksha Chowdhery}, {and} \bibinfo{person}{Denny Zhou}.} \bibinfo{year}{2022}\natexlab{b}.
\newblock \showarticletitle{Self-consistency improves chain of thought reasoning in language models}.
\newblock \bibinfo{journal}{\emph{arXiv preprint arXiv:2203.11171}} (\bibinfo{year}{2022}).
\newblock


\bibitem[Wang et~al\mbox{.}(2023b)]%
        {wang2023rolellm}
\bibfield{author}{\bibinfo{person}{Zekun~Moore Wang}, \bibinfo{person}{Zhongyuan Peng}, \bibinfo{person}{Haoran Que}, \bibinfo{person}{Jiaheng Liu}, \bibinfo{person}{Wangchunshu Zhou}, \bibinfo{person}{Yuhan Wu}, \bibinfo{person}{Hongcheng Guo}, \bibinfo{person}{Ruitong Gan}, \bibinfo{person}{Zehao Ni}, \bibinfo{person}{Man Zhang}, {et~al\mbox{.}}} \bibinfo{year}{2023}\natexlab{b}.
\newblock \showarticletitle{RoleLLM: Benchmarking, Eliciting, and Enhancing Role-Playing Abilities of Large Language Models}.
\newblock \bibinfo{journal}{\emph{arXiv preprint arXiv:2310.00746}} (\bibinfo{year}{2023}).
\newblock


\bibitem[Wei et~al\mbox{.}(2022a)]%
        {wei2022emergent}
\bibfield{author}{\bibinfo{person}{Jason Wei}, \bibinfo{person}{Yi Tay}, \bibinfo{person}{Rishi Bommasani}, \bibinfo{person}{Colin Raffel}, \bibinfo{person}{Barret Zoph}, \bibinfo{person}{Sebastian Borgeaud}, \bibinfo{person}{Dani Yogatama}, \bibinfo{person}{Maarten Bosma}, \bibinfo{person}{Denny Zhou}, \bibinfo{person}{Donald Metzler}, {et~al\mbox{.}}} \bibinfo{year}{2022}\natexlab{a}.
\newblock \showarticletitle{Emergent abilities of large language models}.
\newblock \bibinfo{journal}{\emph{arXiv preprint arXiv:2206.07682}} (\bibinfo{year}{2022}).
\newblock


\bibitem[Wei et~al\mbox{.}(2022b)]%
        {wei2022chain}
\bibfield{author}{\bibinfo{person}{Jason Wei}, \bibinfo{person}{Xuezhi Wang}, \bibinfo{person}{Dale Schuurmans}, \bibinfo{person}{Maarten Bosma}, \bibinfo{person}{Fei Xia}, \bibinfo{person}{Ed Chi}, \bibinfo{person}{Quoc~V Le}, \bibinfo{person}{Denny Zhou}, {et~al\mbox{.}}} \bibinfo{year}{2022}\natexlab{b}.
\newblock \showarticletitle{Chain-of-thought prompting elicits reasoning in large language models}.
\newblock \bibinfo{journal}{\emph{Advances in neural information processing systems}}  \bibinfo{volume}{35} (\bibinfo{year}{2022}), \bibinfo{pages}{24824--24837}.
\newblock


\bibitem[Wu et~al\mbox{.}(2023)]%
        {wu2023deciphering}
\bibfield{author}{\bibinfo{person}{Dekun Wu}, \bibinfo{person}{Haochen Shi}, \bibinfo{person}{Zhiyuan Sun}, {and} \bibinfo{person}{Bang Liu}.} \bibinfo{year}{2023}\natexlab{}.
\newblock \showarticletitle{Deciphering Digital Detectives: Understanding LLM Behaviors and Capabilities in Multi-Agent Mystery Games}.
\newblock \bibinfo{journal}{\emph{arXiv preprint arXiv:2312.00746}} (\bibinfo{year}{2023}).
\newblock


\bibitem[Xi et~al\mbox{.}(2023)]%
        {xi2023rise}
\bibfield{author}{\bibinfo{person}{Zhiheng Xi}, \bibinfo{person}{Wenxiang Chen}, \bibinfo{person}{Xin Guo}, \bibinfo{person}{Wei He}, \bibinfo{person}{Yiwen Ding}, \bibinfo{person}{Boyang Hong}, \bibinfo{person}{Ming Zhang}, \bibinfo{person}{Junzhe Wang}, \bibinfo{person}{Senjie Jin}, \bibinfo{person}{Enyu Zhou}, {et~al\mbox{.}}} \bibinfo{year}{2023}\natexlab{}.
\newblock \showarticletitle{The rise and potential of large language model based agents: A survey}.
\newblock \bibinfo{journal}{\emph{arXiv preprint arXiv:2309.07864}} (\bibinfo{year}{2023}).
\newblock


\bibitem[Xu et~al\mbox{.}(2023)]%
        {xu2023exploring}
\bibfield{author}{\bibinfo{person}{Yuzhuang Xu}, \bibinfo{person}{Shuo Wang}, \bibinfo{person}{Peng Li}, \bibinfo{person}{Fuwen Luo}, \bibinfo{person}{Xiaolong Wang}, \bibinfo{person}{Weidong Liu}, {and} \bibinfo{person}{Yang Liu}.} \bibinfo{year}{2023}\natexlab{}.
\newblock \showarticletitle{Exploring large language models for communication games: An empirical study on werewolf}.
\newblock \bibinfo{journal}{\emph{arXiv preprint arXiv:2309.04658}} (\bibinfo{year}{2023}).
\newblock


\bibitem[Yang et~al\mbox{.}(2024)]%
        {yang2024llm}
\bibfield{author}{\bibinfo{person}{Qisen Yang}, \bibinfo{person}{Zekun Wang}, \bibinfo{person}{Honghui Chen}, \bibinfo{person}{Shenzhi Wang}, \bibinfo{person}{Yifan Pu}, \bibinfo{person}{Xin Gao}, \bibinfo{person}{Wenhao Huang}, \bibinfo{person}{Shiji Song}, {and} \bibinfo{person}{Gao Huang}.} \bibinfo{year}{2024}\natexlab{}.
\newblock \showarticletitle{LLM Agents for Psychology: A Study on Gamified Assessments}.
\newblock \bibinfo{journal}{\emph{arXiv preprint arXiv:2402.12326}} (\bibinfo{year}{2024}).
\newblock


\bibitem[Zeng et~al\mbox{.}(2022)]%
        {zeng2022glm}
\bibfield{author}{\bibinfo{person}{Aohan Zeng}, \bibinfo{person}{Xiao Liu}, \bibinfo{person}{Zhengxiao Du}, \bibinfo{person}{Zihan Wang}, \bibinfo{person}{Hanyu Lai}, \bibinfo{person}{Ming Ding}, \bibinfo{person}{Zhuoyi Yang}, \bibinfo{person}{Yifan Xu}, \bibinfo{person}{Wendi Zheng}, \bibinfo{person}{Xiao Xia}, {et~al\mbox{.}}} \bibinfo{year}{2022}\natexlab{}.
\newblock \showarticletitle{Glm-130b: An open bilingual pre-trained model}.
\newblock \bibinfo{journal}{\emph{arXiv preprint arXiv:2210.02414}} (\bibinfo{year}{2022}).
\newblock


\bibitem[Zhang et~al\mbox{.}(2024)]%
        {zhang2024timearena}
\bibfield{author}{\bibinfo{person}{Yikai Zhang}, \bibinfo{person}{Siyu Yuan}, \bibinfo{person}{Caiyu Hu}, \bibinfo{person}{Kyle Richardson}, \bibinfo{person}{Yanghua Xiao}, {and} \bibinfo{person}{Jiangjie Chen}.} \bibinfo{year}{2024}\natexlab{}.
\newblock \showarticletitle{TimeArena: Shaping Efficient Multitasking Language Agents in a Time-Aware Simulation}.
\newblock \bibinfo{journal}{\emph{arXiv preprint arXiv:2402.05733}} (\bibinfo{year}{2024}).
\newblock


\bibitem[Zhang et~al\mbox{.}(2022)]%
        {zhang2022automatic}
\bibfield{author}{\bibinfo{person}{Zhuosheng Zhang}, \bibinfo{person}{Aston Zhang}, \bibinfo{person}{Mu Li}, {and} \bibinfo{person}{Alex Smola}.} \bibinfo{year}{2022}\natexlab{}.
\newblock \showarticletitle{Automatic chain of thought prompting in large language models}.
\newblock \bibinfo{journal}{\emph{arXiv preprint arXiv:2210.03493}} (\bibinfo{year}{2022}).
\newblock


\bibitem[Zhou et~al\mbox{.}(2024)]%
        {zhou2024lima}
\bibfield{author}{\bibinfo{person}{Chunting Zhou}, \bibinfo{person}{Pengfei Liu}, \bibinfo{person}{Puxin Xu}, \bibinfo{person}{Srinivasan Iyer}, \bibinfo{person}{Jiao Sun}, \bibinfo{person}{Yuning Mao}, \bibinfo{person}{Xuezhe Ma}, \bibinfo{person}{Avia Efrat}, \bibinfo{person}{Ping Yu}, \bibinfo{person}{Lili Yu}, {et~al\mbox{.}}} \bibinfo{year}{2024}\natexlab{}.
\newblock \showarticletitle{Lima: Less is more for alignment}.
\newblock \bibinfo{journal}{\emph{Advances in Neural Information Processing Systems}}  \bibinfo{volume}{36} (\bibinfo{year}{2024}).
\newblock


\bibitem[Zhou et~al\mbox{.}(2023)]%
        {zhou2023webarena}
\bibfield{author}{\bibinfo{person}{Shuyan Zhou}, \bibinfo{person}{Frank~F Xu}, \bibinfo{person}{Hao Zhu}, \bibinfo{person}{Xuhui Zhou}, \bibinfo{person}{Robert Lo}, \bibinfo{person}{Abishek Sridhar}, \bibinfo{person}{Xianyi Cheng}, \bibinfo{person}{Yonatan Bisk}, \bibinfo{person}{Daniel Fried}, \bibinfo{person}{Uri Alon}, {et~al\mbox{.}}} \bibinfo{year}{2023}\natexlab{}.
\newblock \showarticletitle{Webarena: A realistic web environment for building autonomous agents}.
\newblock \bibinfo{journal}{\emph{arXiv preprint arXiv:2307.13854}} (\bibinfo{year}{2023}).
\newblock


\bibitem[Zhouhong~Gu(2023)]%
        {gu2023beyond}
\bibfield{author}{\bibinfo{person}{Jiangjie Chen Haoning Ye Xiaoxuan Zhu Zihan Li Zheyu Ye Yan Gao Yao Hu Yanghua Xiao Hongwei~Feng Zhouhong~Gu, Lin~Zhang}.} \bibinfo{year}{2023}\natexlab{}.
\newblock \showarticletitle{Piecing Together Clues: A Benchmark for Evaluating the Detective Skills of Large Language Models}.
\newblock \bibinfo{journal}{\emph{arXiv preprint arXiv:2307.05113}} (\bibinfo{year}{2023}).
\newblock


\bibitem[Zhu et~al\mbox{.}(2023)]%
        {zhu2023fireball}
\bibfield{author}{\bibinfo{person}{Andrew Zhu}, \bibinfo{person}{Karmanya Aggarwal}, \bibinfo{person}{Alexander Feng}, \bibinfo{person}{Lara~J Martin}, {and} \bibinfo{person}{Chris Callison-Burch}.} \bibinfo{year}{2023}\natexlab{}.
\newblock \showarticletitle{FIREBALL: a dataset of dungeons and dragons actual-play with structured game state information}.
\newblock \bibinfo{journal}{\emph{arXiv preprint arXiv:2305.01528}} (\bibinfo{year}{2023}).
\newblock


\bibitem[Zhuang et~al\mbox{.}(2023)]%
        {zhuang2023efficiently}
\bibfield{author}{\bibinfo{person}{Yan Zhuang}, \bibinfo{person}{Qi Liu}, \bibinfo{person}{Yuting Ning}, \bibinfo{person}{Weizhe Huang}, \bibinfo{person}{Rui Lv}, \bibinfo{person}{Zhenya Huang}, \bibinfo{person}{Guanhao Zhao}, \bibinfo{person}{Zheng Zhang}, \bibinfo{person}{Qingyang Mao}, \bibinfo{person}{Shijin Wang}, {et~al\mbox{.}}} \bibinfo{year}{2023}\natexlab{}.
\newblock \showarticletitle{Efficiently measuring the cognitive ability of llms: An adaptive testing perspective}.
\newblock \bibinfo{journal}{\emph{arXiv preprint arXiv:2306.10512}} (\bibinfo{year}{2023}).
\newblock


\end{thebibliography}
\appendix
\end{CJK}
\end{document}